# Restoring Arabic vowels
# through omission-tolerant dictionary lookup

تشْكيل الكَلِمات عَبْرَ مَوارد حاسوبيّة

*Alexis Amid Neme and Sébastien Paumier*
Université Paris-Est, LIGM, UPEM, CNRS, ENPC, ESIEE, 77454, Marne-la-Vallée, France

Abstract

Vowels in Arabic are optional orthographic symbols written as diacritics above or below letters. In Arabic texts, typically more than 97 percent of written words do not explicitly show any of the vowels they contain; that is to say, depending on the author, genre and field, less than 3 percent of words include any explicit vowel. Although numerous studies have been published on the issue of restoring the omitted vowels in speech technologies, little attention has been given to this problem in papers dedicated to written Arabic technologies.

In this research, we present Arabic-Unitex, an Arabic Language Resource, with emphasis on vowel representation and encoding. Specifically, we present two dozens of rules formalizing a detailed description of vowel omission in written text. They are typographical rules integrated into large-coverage resources for morphological annotation. For restoring vowels, our resources are capable of identifying words in which the vowels are not shown, as well as words in which the vowels are partially or fully included. By taking into account these rules, our resources are able to compute and restore for each word form a list of compatible fully vowelized candidates through omission-tolerant dictionary lookup.

In our previous studies, we have proposed a straightforward encoding of taxonomy for verbs (Neme, 2011) and broken plurals (Neme & Laporte, 2013). While traditional morphology is based on derivational rules, our description is based on inflectional ones. The breakthrough lies in the reversal of the traditional root-and-pattern Semitic model into pattern-and-root, giving precedence to patterns over roots.

The lexicon is built and updated manually and contains 76,000 fully vowelized lemmas. It is then inflected by means of finite-state transducers (FSTs), generating 6 million forms. The coverage of these inflected forms is extended by formalized grammars, which accurately describe agglutinations around a core verb, noun, adjective or preposition.

A laptop needs one minute to generate the 6 million inflected forms in a 340-Megabyte flat file, which is compressed in two minutes into 11 Megabytes for fast retrieval. Our program performs the analysis of 5,000 words/second for running text (20 pages/second).

Based on these comprehensive linguistic resources, we created a spell checker that detects any invalid/misplaced vowel in a fully or partially vowelized form. Finally, our resources provide a lexical coverage of more than 99 percent of the words used in popular newspapers, and restore vowels in words (out of context) simply and efficiently.

# Abstract in Arabic

الحركاتُ رموزٌ إختياريٌ كتابتها في اللّغة العربية، وتُكتب كل حركةٍ فوق أو تحت الحرف المُناطة إليه. تشمل معظم النصوص العربية على كلمات مُشكلة جزئيًا ولا يتعدى عامةً نسبتها 3٪ من الكلمات وهذه النسبة تتوقف على الناشر والكاتب والميدان المتخصِص. على الرغم من أن العديد من الدراسات العلمية قد تمّ نشرها في مسألة حذف الحركات في تقنيات الكلام، فقد أوليَ اهتمام لا يذكر لنفس المشكلة في الدراسات المخصصة لتقنيات العربية المكتوبة.

في هذا البحث، نقدم وصفاً مفصلاً لحذف الحركات في النصوص المكتوبة والقواعد المطبعية ذات الصلة وقواعد حذفهما في الموارد الحاسوبيَّة. مواردنا قادرة على التعرُّف على الكلمات المشكلة كلياً أو جزئيا أو غير المشكلة كما وإعادة الحركات لكلٍ منها،

في دراسات سابقة، اقترحنا تصنيفات للأفعال (Neme, 20011) وتصنيفات لجموع لتكسير (Neme & Laporte, 2013) مبنية على أُسُس الصرف التقليدي. ففي حين يحتوي علم الصرف التقليدي على توصيف القواعد الاشتقاقية وغير الاشتقاقية، يستند وصفنا على الصرف غير الإشتقاقي حصراً. والجديد في مقاربتنا يكمن في عكس مقاربة علم الصرف التقليديّة التي هي معادلة (الجذر-الوزن) إلى (الوزن-الجذر) مع إعطاء الأولوية للوزن على حساب الجذر. هذا التغيير سمح لنا التعرّف على الفعل كمدخل معجمي بشكل أسرع وأدقّ وبالتالي التعرّف على جذره ووزنه، كما قلّص تحديد وبرمجة مئات القواعد الصرفيّة والإملائية التي تربط أشكال الفعل بجذره ووزنه.

وقد تم بناء المورد اللغوي يدويًا ويحتوي على 76000 مدخل معجمي محرّك بأكمله. تمّ تصريف هذا المورد ليحتوي على 6 ملايين شكل محرّك أيضاً. وقد تمّ إضافة السوابق واللواحق لهذه الأشكال عن طريق قواعد تلاصقيّة نحويّة دقيقة حول فعل أساس، إسم، أو صفة. هذه القواعد تحدِّد تتابع الشرائح المسموح بها من سوابق ولواحق حول شريحة أساسية.

يحتاج حاسوب محمول إلى دقيقة واحدة لتوليد 6 ملايين شكل محرّك وحجم الملف 340 ميغابايت، قد تمّ ضغطه إلى 11 ميغابايت للبحث السريع. يقوم برنامجنا بتحليل 5000 كلمة في الثانية (20 صفحات/ثانية). والتغطية المعجمية لمواردنا تضاهي 99٪ لنصوص من الصحف العامة.

في هذه الدراسة، نركز على توصيف قواعد حذف الحركات والشدّة والهمزة. ونعرض حلاُ بسيطاً فعالاً وأنيقاً يتعرَّف على كلمات غير مشكلة أو مشكلة جزئياً أو كلياً وإعادة الحركات لكلٍ منها في برنامج للتحليل الصرفي.

## 1  Introduction

Writing conventions in Arabic are characterized by being based on consonants and also underspecified—they usually lack short vowels and other diacritics. This is indirectly connected to the historical legacy of the first consonantal Phoenician alphabet, as is the case with other Semitic languages. In practice, speakers and readers do restore these essential lacking pieces based on their memory and knowledge of Arabic. Therefore, it is a legitimate goal that computers should be able to compute and restore these missing vowels and diacritics in written texts.

Big institutions were unsuccessful in dealing with the issue of missing vowels in written texts. Googlelabs withdrew its software to restore vowels in Arabic text in 2012, just a year after its release, while in May 2012 an Arabic spell checker for Gmail was released only to be withdrawn the same year. One of the problems users encountered using Gmail's spell checker was that it erroneously flagged as mistakes fully or partially vowelized words which happened to be correct. Microsoft Office 2016 suffers from the opposite problem: its Arabic spell checker ignores fully or partially vowelized words - erroneous vowels are not flagged as mistakes and neither are typographical mistakes such as the *'-bF'* and *'-AN'* endings in كتابً كتاباً *ktAbF* or *ktAbAN*.[1]

Lately, maybe in 2016, Google released an Arabic spell checker with a low coverage of inflection and of affixed and agglutinated words. This time, like Microsoft, it ignores partially vowelized words; even worse, it does not flag a wrong word if it contains one vowel. In average, Google's spell checker flags around 10% of valid words erroneously.

These problems highlight the difficulties in building accurate Arabic computational and morphological resources. There are a number of reasons for this:

- Arabic has a rich morphology, containing six attributes for verbs and four for nouns and adjectives
- its inflection uses prefixes, suffixes, and mostly infixes described by the root-and-pattern traditional model
- words may have agglutinated clitics (from a set of around 30 clitics)
- vowels in words are generally omitted or partially represented.

If the first three issues have been handled in Arabic Language Technologies with some degree of attention, the last issue is less studied in computational morphology and has not been given the correct rank of importance, as Maamouri et al. (2006) state[2]: *In general, the role of diacritics in a NLP pipeline that includes parsing is very much an open question.*

---

[1] The TB++ transliteration used in this paper is derived from the Buckwalter encoding and adopted in Unitex to map Arabic <=> Latin: ء, c; آ, C; أ, O; ؤ, W; إ, I; ئ, e; ا, A; ب, B; ة, p; ت, T; ث, V; ج, J; ح, H; خ, x; د, d; ذ, J; ر, r; ز, z; س, s; ش, M; ص, S; ض, D; ط, T; ظ, Z; ع, E; غ, g; ف, f; ق, q; ك, k; ل, l; م, m; ن, n; ه, h; و, w; ى, Y; ي, y; ً, F; ٌ, N; ٍ, K; َ, a; ُ, u; ِ, i; ّ, G; ْ, o.

[2] There are optional typographical signs in another Semitic language. "The Hebrew script [has two variants]: one in which vocalization diacritics, known as niqqud "dots", decorate the words, and another in which the dots are missing, and other characters represent some, but not all of the vowels. Most of the texts in Hebrew are of the latter kind; unfortunately, different authors use different conventions for the undotted script. Thus, the same word can be written in more than one way, sometimes even within the same document, again adding to the ambiguity." (Wintner, 2008)

Many Arabic lexical resources lack information about vowels, an absence often explained by the rarity of vowels in written texts. This is a view that is becoming widespread with the expansion of corpus linguistics.

However, spelling out vowels in words is a convenient way to distinguish lemmas with different meanings: *Eaqod/Eiqod/Eaqid* "contract/necklace/thickening (for a liquid)" عَقْد/عِقْد/عَقَد or *giloyaAn/galayaAn* "is boiling (adjective)/the boiling (noun)" غَلَيان غِلْيان. Vowels and other diacritics are part of the message, even if they are not represented as graphical symbols. Language is foremost an oral form of communication and the selection of writing conventions is subsequent. Vowels are an essential part of Arabic, even if they lack in its written form. Why would such an essential part of the language be irrelevant to NLP, or less relevant than POS?

Creating Arabic lexical resources is not a simple task. Making them accurate without vowels is impossible. For example, in some words, the short vowel after the first consonant alternates with a variant: *nufaAyap* vs. *nifaAyap* "rubbish" (whereas *\*nafaAyap* is inacceptable), and the prevalence of a choice in a text may indicate a regional pronunciation or a register of language: formal or colloquial. In all Arabic dictionaries, both old and modern, diacritical information is available and inventoried thoroughly. For speech technologies, vowels are required.

By 'accurate' ALR, we mean both recall (high lexical coverage) and precision (rejection of invalid forms), at three levels:

- <u>inflection</u>: if a verb or noun is in the ALR, then all the inflected forms of its lemma and no invalid inflected forms must be taken into account;

- <u>agglutination</u>: if an inflected form is in the ALR, then all of its valid agglutinated forms, and no invalid forms, must be taken into account;

- <u>vowelization</u>: if an inflected form, agglutinated or not, is in the ALR, then all of its vowelized forms, whether it is partial or total vowelization, must be taken into account, as well as forms not containing vowels, and no invalid forms. [3]

Devices (involving programs, extensive lists, FSTs, etc.) recognizing and/or generating such forms should not over- or under-generate.

The orthographic system of Arabic includes 34 'bare letters', which are always transcribed, and nine diacritical marks optionally written:
- Three short vowels (*a, i, u*) and the zero-vowel diacritic or *sukoon* (*o*), for the absence of a vowel; all four occur in all positions except word-initial, although *o* occurs very rarely between the first and second consonants;
- Three nunation marks (*-N, -F, -K*, phonetically equivalent to *-un, -an, -in*) used as noun case and definiteness (indefinite) suffixes, and therefore only in ending positions;
- the gemination mark ّ or *shadda* (*G*), which is used for the derivation of new words or broken plural inflection and occurs after the second consonant of the main morphological element of the word;
- the superscript long 'a' or superscript *alif* ٰ (*R*), a rarely scripted, archaic form usable in some frequent words such as the pronoun هذا *haRJaA* 'this' and in some archaic spellings still used in modern Arabic such as *raHomaRn* 'merciful'.

---

[3] One may add a typographical consistency at document(s) level, in terms of the so-called editing style of a publication. In French for instance, this requirement includes using the same symbol in words such as *oeuvre* or *œuvre* throughout one or a set of documents; in Arabic, it will be a mandatory transcription of a hamza-above-alif.

Moreover, an initial glottal stop or *hamza* can be omitted. In a word initial position, it is represented by two characters: *O* for *hamza* above *A* أ; and *I* for *hamza* under *A* إ. Omitting the glottal stop consists of writing the *A* ا instead of *O* or *I*; therefore, these two characters belong to our topic in this paper. In non-initial position, the *hamza* diacritic appears in five different characters (*c,* ء; *W,* ؤ ; *e,* ئ; *O,* أ; *I,* إ); but it is not an optional diacritic and cannot be omitted. Consequently, these characters with *hamza* in non-initial position do not belong to our topic.

For simplicity, we use interchangeably 'vowel' and 'diacritic' throughout the rest of the paper and we mean by both terms all nine diacritical marks, and the initial *hamza* diacritics carried by *A*.

Diacritization/vowelization is the operation to assign/restore a diacritic/vowel to a undiacritized/unvowelized consonant in a word. It is a typical knowledge test in Arabic vocabulary and grammar. Words with at least one written vowel are said to be partially vowelized; and fully vowelized, when all are written. A word form delimited by two spaces may include one or two vowels (three in rare cases). "*In the Penn Arabic Treebank (part 3), 1.6 percent of all words have at least one diacritic indicated by their author*" (Habash, 2010, p.11). In most newspapers, only about 2-3 percent of words are partially vowelized, although this can reach 12-15 percent in well-edited articles. Some reference books are almost completely vowelized, such as *Kitab fasl al-maqal* by Averroes, the Andalusian philosopher of the XII[th] century; while other books including dictionaries, teaching textbooks and holy texts are fully vowelized.

"*Arabic NLP research faces two major challenges, not necessarily shared with many other natural languages: the first is its complex linguistic structure and the second, the specific features of its orthographic system*" (Maamouri et al. 2006, Introduction). In the next sub-section, we present the main consequence of under-representation of vowels on morphological analysis: it increases tagging ambiguities.

DIACRITICAL AMBIGUITY

Word-level ambiguity is common to all natural languages, including Arabic; even the full representation of vowels does not prevent ambiguity in Arabic, as in *EaAmil*, "worker/agent" عامل. However, the under-specification of Arabic script – the loss of vowels – causes written Arabic to have more ambiguities, called *diacritical ambiguities*. We restrict the definition of diacritical ambiguity to the case where the omission of one or more vowels generates additional ambiguity.

To illustrate diacritical ambiguity in Arabic, let us draw a parallel with French examples with or without accent(s). In French, poor and rich typography refers respectively to non-accented and accented typography. In order to make a parallel with vowel omission in Arabic, we extend the use of the term 'poor typography' to the case where at least one accent is omitted, and at least another is present. The rich word form *chantées* has only one possible poor typographical representation *chantees,* whereas *déjà* also has two possible partial accentuations *déja* and *dejà*. A word form such as *déjà* has four possible typographical representations: fully, partially accentuated or not accentuated.

How to retrieve the fully vowelized form from a partially vowelized one? An index is the simplest way to access stored information through a keyword. Thus, in order to access a fully accented word in a French lexicon, one may build an auxiliary index on the `poor:rich` pattern by replacing each accented letter by its non-accented counterpart:

```
chantees:chantées
chantées:chantées
deja:déjà
déja:déjà
dèjà:déjà
déjà:déjà
```

Conversely, the form *chantéés* would be inexistent in such an index since only the omission of an accent is valid, not the addition (as in *katab**aa**tu* in Arabic); the form *chanteees* would also be inexistent, since it has no corresponding valid rich form.

If an index for word forms like *chantées* is simple to construct, the index for *déjà* exhibits more complexity. Arabic word forms are more complex than *déjà* because in the full representation of a word form, a diacritic occurs after each consonant. Building such an index for Arabic would not be a viable solution because it would contain several billions of partially vowelized forms.

There is no diacritical ambiguity in the words *deja* and *déja* since they refer to a single fully accented form: *déjà*. A complex diacritical ambiguity would be the poor typographical representation of *pêche, péché, pèche, péché, pêche, pêché,* (resp. "peach", "sin", "(he) sins", "sinned", "fishing", "fished"). All six are represented in poor typography by *peche*. So, the under-representation of accents in *peche* is the origin of an ambiguity between 6 candidates. But partial representation of diacritics, as in *pêche*, reduces them from six to three. It is a pity not to take advantage of such information in a parser (cf. Sections 2.3 and 2.4).

Serbian exhibits similar features; only 5% of words in ordinary text contain at least one accentuated letter, and many of them have no diacritical ambiguity since they stand for a single fully accented form like *déjà*. In "*Knowledge and Rule-Based Diacritic Restoration in Serbian*", Krstev et al. (2018) propose a solution for Serbian based entirely on linguistic resources. They present *"a procedure for the restoration of diacritics in Serbian texts written using the degraded Latin alphabet. The procedure relies on the comprehensive lexical resources for Serbian: the morphological electronic dictionaries, the Corpus of Contemporary Serbian (processed for uni-, bi- and tri-gram frequencies) and local grammars. Dictionaries are used to identify (in 5 modular steps) possible candidates for the restoration, while the data obtained from SrpKor and local grammars assists in making a decision (defined by 7 steps) between several candidates in cases of ambiguity"*. They conclude, *"The diacritic restoration can be successfully solved by using a rule-based approach that relies on the lexical resources. […]. This solution exhibits the advantage of transparency (and modularity) which is usually characteristic of such methods."*

Fig. 1 illustrates partial diacritization with some statistical data about a 200-word excerpt of a newspaper text about "the rising price of gold".

لا تقتصر أهمية الذهب وقيمته على كونه أداة للتزيّن، بل على الدور الذي يضطلع به في تهدئة مخاوف المستثمرين في الأيام الأكثر تشاؤماً، باعتباره ملاذاً آمناً يقيهم شر التراجع في الأسواق المالية. وفي ظل تضافر العوامل التي يمكن أن تضغط على أسعاره صعوداً أو هبوطاً، يبقى مصير المعدن الأصفر رهن التطورات المقبلة، تماشياً مع تراجع الاقتصاد الصيني والتوقعات باحتمال رفع الفيدرالي الأميركي أسعار الفائدة مرتين هذه السنة.

> وتجتمع عوامل عدّة للتأثير إيجاباً على أسعار الذهب، كالتحفيز النقدي مثلاً الذي يعتمده حالياً كلٌ من المصرف المركزي الأوروبي والمصرف المركزي الصيني، وتراجع أسعار النفط الذي ينشط عملية اللجوء إلى الملاذات الآمنة، لأنّه يحضّ المستثمرين في العقود الآجلة للنفط على وضع حد لهذا الإستثمار، فضلاً عن أنَّ ارتفاع نسب الفوائد في الولايات المتحدة الأميركية وتالياً تعزيز الطلب على الدولار، يدفع أسعار المعدن الأصفر نزولاً. وفي هذا السياق، يشير بو سليمان إلى أنَّ حالات التضخم كفيلة هي أيضاً برفع أسعار الذهب، في حين أنَّ الانكماش يهبط بها. أخيراً، يتوقع بو سليمان أن تراوح أسعار الذهب خلال السنة الجارية بين 950 و1200 دولار نتيجة التضارب الحاصل في الأسواق العالمية، إذ إنَّ الإقتصاد الأميركي أظهر بوادر تعاف، في حين أنَّ نظيره الصيني تراجع، بما يضمن عدم سلك الأسعار مساراً انحدارياً وتالياً تحقيق التوازن في السوق .
>
> Diacritics are included by authors to facilitate reading.
>
> Among the 404 words, 50 (in red above) are partially vowelized: 38 with one diacritic and 12 with two vowels. The 50 diacritics are: 26 -*AF,* 23 *G,* 10 *a,* 2 *u,* 1 -*N.*
>
> In *Annahar* (Beirut) and *Al-Hayat* (Saudi Arabia), which are reference newspapers in Arab countries, the percentage of partially vowelized words is often estimated to 2-3 percent[4], but this rate also depends on the journalist and the field, as articles on special topics tend to include more diacritics.
>
> The -*AF* ending is used to mark the accusative or the adverbial POS that may be confused with the dual if the *F* is omitted.
>
> The -*Ga-* sequence is often used to disambiguate between conjunctions: *InGa, OnGa, Ono*
>
> The -*G-* gemination diacritic is often used in 2 or 3-letter words, such as in quantifiers or bi-literal verbs, but also to avoid confusion between simple tri-literal and derived tri-literal verbs.

Fig. 1. An extract from *Annahar* of 13 January 2016 with partial vowelization (http://www.annahar.com/article/301388)

In Section 2, we present previous work about building ALR and the (un)reliability of these resources for diacritic restoration. In Section 3, we make a general presentation of Arabic-Unitex as a full-form diacritized ALR. In Section 4, we detail our solutions in Arabic-Unitex for diacritic omission rules and related typographical issues. In section 5, we present the Arabic-Unitex tagset, lexicon figures and performance. In section 6, we detail our compression algorithm for Semitic languages and our algorithm for restoring Arabic vowels for words (out of context) through omission-tolerant dictionary lookup.

## 2   Previous Work

Studies focusing on diacritics in Arabic Speech Technologies, and especially in Text-to-Speech (TTS), are numerous since restoring omitted vowels is critical for syllabification. TTS systems *inevitably* contain such functionality for restoring vowels; whereas this functionality is *optionally* included in systems processing written text. Zitouni et al. (2006)[5] report Word Error

---

[4] According to our corpus study of 6930 words from the *Annahar* newspaper, 209 words (3%) include at least a diacritic (Neme, 2011, Section 4.2).

[5] "The lack of diacritics may lead to considerable lexical ambiguity that must be resolved by contextual information, which in turn presupposes knowledge of the language. It was observed in (Debili et al., 2002) that a

Rates (WER) in diacritization ranging from 10 percent for lexical diacritics to 25 percent where case endings are included.

Contrariwise, in Arabic Natural Language Processing, few papers are dedicated to Arabic vowelization, *"still largely understudied in the current NLP literature"* (Maamouri et al , 2006). There are many reasons: *"since non-diacritized text prevails, the Arabic NLP community seems to have accepted using it as the de facto 'real world' information material without feeling an obligation to question its choice/use, even espousing the idea sometimes that the robustness of software algorithms can deal with the problem and reduce the negative effect of the missing information on their research."* […] *"The prohibitive cost and the usually unequal and questionable quality of human/manual diacritization have led the scientific Arabic NLP community and its sponsors to focus more on volume of unvowelized data so far"* (Maamouri et al , 2006).

One may wonder if Arabic Speech Technologies Speech-To-Text (STT) and Text-To-Speech (TTS) approaches to diacritization might be adapted to written text technologies. But TTS and written text processing approaches to restoring diacritics use similar techniques: rule-based, statistical, and hybrid approaches; and they face the same challenges: sparseness of data since Arabic is morphologically rich and agglutinated, Out-Of-Vocabulary tokens, scarcity of modern Arabic vowelized resources, etc. Thus, there is no reason to speculate that adaptation of current TTS technologies might bring about any key innovation in diacritization of written text.

Alternatively, STT might be used to overcome the present scarcity of diacritized corpora in Modern Standard Arabic, by implementing an ambitious programme of accurate transcription of audio recordings of formal news. However, such an undertaking would involve post-edition, and even with massive investment, would probably not remedy more than *partially* the lack of training data. Therefore, the availability of more training data will not dispense from exploiting large coverage lexicon and accurate grammatical rules. *"Hybrid approaches in many surveyed systems perform better as these techniques are guided by language-dependent rules […] Inflection property of Arabic may cause many words to be unseen in learning phase.[…] Pure statistical approaches usually give unsatisfactory performance with unseen data, especially in complex languages that suffer from sparseness as is the case with Arabic, a highly inflected language. This sparseness may cause training data to be insufficient."*.(Azmi and Almajed, 2015, Section 5)

## 2.1 ARACOMLEX (2006-2015)

Not only have commercial packages failed in handling vowels but also research groups have omitted vowels in ALR, such as AraComLex 1.0. *"The decision to ignore diacritics was taken after examining a corpus of 4.5 million Arabic words, where only 54 (sic) words were found to carry meaningful diacritic marks, which is statistically insignificant."* (Attia A. Mohammed, 2006).

---

non-diacritized dictionary word form has 2.9 possible diacritized forms on average and that an Arabic text containing 23,000 word forms showed an average ratio of 1:11.6."

In this sub-section, we discuss the extended version of *AraComLex* (Attia et al., 2011, 2015) because of its representativeness: recently created, available publicly, well documented and based on a sound methodology, it may be considered to represent the current state of the art in the domain of ALR; and a new trend attempting to build a full coverage of an ALR. We also mention some resources derived from *AraComLex*.

AraComLex 1.0 (Attia, 2006) has 10,800 lemmas; Attia et al. (2011) have increased semi-automatically their resource to reach 30,587 lemmas, arguing that creating a lexicon is time-consuming: *"Creating a lexicon is usually a labour-intensive task. For instance, Attia took **three years** in the development of his morphology, while SAMA and its predecessor, Buckwalter's morphology, were developed over **more than a decade**, and at least seven people were involved in updating and maintaining the morphology. […] and [we have built] a large-scale open-source finite-state morphological transducer for Arabic, AraComLex, that contains 30,587 lemmas. AraComLex generates 12,951,042 words."* According to the authors, the lexical coverage rate for general news or semi-literary text is around 86%. They add, *"The quality and coverage of the lexical database determines the quality and coverage of the morphological analyser, and limitations in the lexicon will cascade through to higher levels of processing […]"*

A common method to create a reliable reference list of words for a language is inspired from corpus linguistics: it consists in collecting corpora of several gigabytes, removing duplicate words, and validating the unique words semi-automatically. But, as Attia et al. (2015) notice: *"due to the richness and complexity of Arabic morphology, there is no corpus, no matter how large, that contains all possible word forms. Given a word in Arabic, one can change its form by adding or removing yet another prefix, suffix, proclitic or enclitic. This is why a morphological generator is essential in creating an adequate list of possible words."* (Attia et al., 2015).

Generation of word forms with affixes and clitics is required, indeed. However, it does not resolve another shortcoming of the corpus-based approach: this approach limits the coverage of the dictionary to that of the corpus.

## 2.2 BAMA (2002)

The well-known Buckwalter Arabic Morphological Analyzer (BAMA) is one of the best Arabic morphological analyzers and is available as open source. The BAMA lexicon is considered the baseline of Arabic computational processing. The BAMA uses a concatenative lexicon-driven approach (Buckwalter, 2002) based on three lexica, labelled A, B and C, where B is a multi-stem lexicon, and on a lookup algorithm based on compatibility constraints within the string ABC. In order to match a surface form, the parsing algorithm uses the lexicon's unvowelized stem field and the corresponding ad-hoc category provided in the lexicon: it selects compatible (proclitics and) prefixes and suffixes (and enclitics) in two precompiled lists (cf. Neme, 2011, Section 2).

Buckwalter (2007, 3.6) explains the advantage of BAMA (2004) compared to the Beesley-Xerox solution (Beesley, 1989-2001). The latter is an intricate solution based on twelve lexica, the traditional root-and-pattern model, two-level FST morphology, a large pool of rules formalized to be used with XFST and a lookup algorithm slowed down mainly by the pool of rules. We do agree on Buckwalter's critics to the Beesley-Xerox solution. Even with an important team and support, it is not viable (see Neme & Laporte, 2013 section 2, and "*On the Misuse of Finite State Technology in Semitic Languages: Hebrew and Arabic*", 30 pages, to be published).

The Buckwalter stem-lexicon is constituted by 92,814 stem lines representing 41,178 lemmas, which amounts to a ratio of 2.27 stem/lemma. As an example, Table 2.2 shows the encoding of the lemma '*>aSiyl*' 'authentic' أصيل with its broken plural which admits three orthographic variants determined by case and agglutinated enclitics: '*>uSalA&-u_hu*' (nominative) أصلاؤه , '*>uSalA'-a_hu*' (accusative) أصلاءه , '*>uSalA&-i_hi*' (genitive) أصلائه . Inflectional attributes values are assigned through values attached to affixes.

*Table 2.2. Stem-based representation of the adjective >aSiyl in the BAMA lexicon*

| LEMMA_ID | unvowelized stem | vowelized stem | Morphological category | parser Output form/POS | Line # | Feature values | initial grapheme alif | in Arabic |
|---|---|---|---|---|---|---|---|---|
| >aSiyl_1 | **>Syl** | >aSiyl | **N/ap** | **>aSiyl/ADJ** | 1 | sing; sing+pro | hamza-above (O) | أصيل |
| >aSiyl_1 | **ASyl** | >aSiyl | **N/ap** | **>aSiyl/ADJ** | 2 | sing; sing+pro | bare-alif (A) | اصيل |
| >aSiyl_1 | **>SlA'** | >uSalA' | **Ndip** | **>uSalA'/ADJ** | 3 | plu; plu-acc+pro | O | أصلاء |
| >aSiyl_1 | **ASlA'** | >uSalA' | **Ndip** | **>uSalA'/ADJ** | 4 | plu; plu-acc+pro | A | اصلاء |
| >aSiyl_1 | **>SlA&** | >uSalA& | **Nuh** | **>uSalA&/ADJ** | 5 | plu-nom+pro | O | أصلاؤه |
| >aSiyl_1 | **ASlA&** | >uSalA& | **Nuh** | **>uSalA&/ADJ** | 6 | plu-nom+pro | A | اصلاؤه |
| >aSiyl_1 | **>SlA}** | >uSalA} | **Nihy** | **>uSalA}/ADJ** | 7 | plu-gen+pro | O | أصلائه |
| >aSiyl_1 | **ASlA}** | >uSalA} | **Nihy** | **>uSalA}/ADJ** | 8 | plu-gen+pro | A | اصلائه |

In Table 2.2, only the fields in **bold** are used directly by the BAMA parser, the other fields are for managing the lexicon and the last three columns are notes by the authors of this paper. The +pro feature indicates a variant with a mandatory pronoun and its absence a form used without a pronoun: the third and fourth lines represent variants in the plural without pronoun in whatever case, or in the accusative with a pronoun. Note the redundancy between unvowelized/vowelized stem fields. There are duplicates, for example the fifth and sixth lines: both of them represent plural nominative forms with a mandatory pronoun, the only difference being the omission (*A*, bare-alif) or not (*O*, hamza-above) of the initial glottal stop.

In the stem-based approach to the lexicon, a noun with broken plural (BP) and ending glottal stop normally requires four stem forms: one for the singular form and three for the BP. The three BP forms are the stem variants depending on the noun case and the occurrence of a pronoun. But since the word '>aSiyl' may begin either with *bare-alif 'A'* or with *alif-with-hamza-above '>'*, it requires a duplication of stems in the lexicon, i.e. four more stem entries are necessary to handle the possible orthographies[6].

We have calculated the number of cases of initial *alif* spelling variation which require stem duplications in BAMA, which is the number of orthographic stem duplications related to an initial *O* (*alif-with-hamza-above*) or *I* (*below*) with the *A* (*bare-alif*) variant. The amount of added stems is 12,204 stems out of 92,814 (13%). This solution for initial glottal-stop diacritics is unsatisfactory. The redundancy of these additional stem-entries and of other duplicated fields (vowelized/unvowelized stem) is error-prone, and very unnatural to Arabic linguists, making the maintenance of the dictionary unnecessarily tricky. Duplication of entries in a manually maintained dictionary has the same drawbacks as code duplication in software engineering: it duplicates the effort required to detect errors, correct them and construct new items.

## 2.3 MADA (2007) and partial diacritization

Hamdi A. (2012) notes that almost all the morpho-syntactic taggers such as Buckwalter (Buckwalter, 2004), Xerox (Beesley, 2005) or MADA (Habash and Rambow, 2007) take as input texts with words partially diacritized, and remove all diacritics, and therefore do not exploit diacritics to disambiguate words. He implements for the MADA analyser (see Table 2.3.b) a solution which takes into account partial vowelization by excluding candidate analyses. The solution is built on the incompatibility between the partially vowelized surface forms and their lexical representation by means of the intersection of two Finite-State-Automata.

To assess performance, Hamdi A. (2012) uses six test sets derived from a single corpus of 25,000 words. The six test sets (in Table 2.3.b) differ as regards the percentage of partially vowelized words: 0%, 1.3%, 10%, 40%, 70% and 100%. The set with 1.3 percent of words is the original corpus, partially vowelized naturally by its authors; the set with 100% is a fully vowelized version, created manually; the other three partially vowelized sets are generated randomly from the fully vowelized set. The baseline of MADA, on the artificially de-vowelized set, is 84.25 percent (Table 2.3.a) of correct morphological analysis. On the set with 1.3 percent of vowelized words, the analysis improves to 84.91 percent. The improvement by 0.66 percentage point reflects the authors' intuitive partial vowelizing of difficult words to make reading easier.

Table 2.3.a. *MADA performance on a corpus of 25K words (from Hamdi, 2012)*

| Criteria | Diacritization | Grammatical tagging | Morph. Analysis |
|---|---|---|---|
| Performance (read Accuracy) | **86.38%** | 96.09% | 84.25% |

---

[6] In the HAMSAH Hebrew project (Wintner, 2008), an XML encoded lexicon, similar redundancies are observed: dotted/undotted. An example with the lexical entry of *bli* "without":

```
<item id="4917" translit="bli" dotted="xxd" undotted="xxu">
    <conjunction type="coord"/>
</item>
```

*Table 2.3.b. Performance of MADA taking into account diacritics (from Hamdi, 2012)*

| Diacritization Rate | MADA Performances | | |
|---|---|---|---|
| | Diacritization | Grammatical tagging | Morph. Analysis |
| 1.3% | **86.97%** | 96.41% | 84.91% |
| 10% | 88.47% | 96.79% | 86.28% |
| 40% | 91.74% | 97.12% | 89.48% |
| 70% | 94.85% | 97.33% | 92.51% |
| 100% | 98.01% | 97.49% | 95.59% |

The MADA research group also created the MAGEAD system (Habash, Rambow, 2006; Altantawy *et al.*, 2010, 2011), implemented with FST technologies and a formalism that mixes inflexional classes and rule-based morphology.

The MAGEAD lexical data are borrowed from Buckwalter (2002): 8 960 verbs (Altantawy *et al.*, 2011:122) and 32 000 nouns and adjectives, admitting broken and suffixed plural (Altantawy *et al.*, 2010:854), but the coverage of broken plural nouns includes only a formalization of triliteral entries: 'we are not evaluating our lexicon coverage (…) Our evaluation aims at measuring performance on words which are in our lexicon, not the lexicon itself. Future work will address the crucial issue of creating and evaluating a comprehensive lexicon' (Altantawy *et al.,* 2010:856; see Neme & Laporte, 2013, Section 2.4.2, for more details). MAGEAD project's latest publication was in 2011.

## 2.4 MADAMIRA (2014)

MADA uses the BAMA lexicon and is based on the native algorithm of BAMA written in PERL. MADAMIRA (2014) is a new version of MADA also offering a coverage of the Egyptian dialect, and implemented in Java: "MADAMIRA follows the same general design as MADA with some additional components inspired from AMIRA"; it is thus "a system for morphological analysis and disambiguation of Arabic that combines some of the best aspects of two previously commonly used systems for Arabic processing". MADAMIRA is "implemented in Java, which provides substantially greater speed than Perl and allows new features to be quickly integrated with the existing code." The reference to Perl alludes to the lexicon and algorithm of BAMA (2002): any implementation using the BAMA lexicon is dependent of the BAMA native algorithm, so MADAMIRA had to reimplement this algorithm in Java.

MADAMIRA uses SAMA 3.1 (2010, https://catalog.ldc.upenn.edu/LDC2010L01), an enhanced version of BAMA involved in the Arabic Treebank. Proclitics/prefixes and suffixes/enclitics in SAMA were extended compared to BAMA, but the lexical coverage remains almost the same with lemmas, instead of the 38,600 lemmas in BAMA (2002). The goal of MADAMIRA is apparently the implementation with Java of the disambiguation with statistical approaches.

Table 2.4.a Evaluation of MADAMIRA accuracy (From Table 3, MADAMIRA, 2014)

| Evaluation Metric | MADA | MADAMIRA | NOTES |
|---|---|---|---|
| EVALDIAC | 86.4 | 86.3 | EVALDIAC: Percentage of words where the analysis chosen by MADAMIRA has the correct fully diacritized form and an exact spelling |
| EVALLEX | 96.2 | 96.0 | EVALLEX: Percentage of words where the chosen analysis has the correct lemma |
| EVALPOS | 96.1 | 95.9 | EVALPOS: Percentage of words where the chosen analysis has the correct part-of-speech |
| EVALFULL | 84.3 | 84.1 | EVALFULL: Percentage of words where the analysis chosen by MADAMIRA has all the features above [EVALDIAC + EVALLEX + EVALPOS]. |

In all metric aspects, MADAMIRA represents a deterioration of accuracy compared to MADA for Standard Arabic. Moreover, MADAMIRA does not take into account Hamdi's critics of MADA (2005).

Fig. 2.4. Screenshot of MADAMIRA with an input sentence (translation in English: *That difference, a small vowel makes it happen in the meaning of words such as 'of ligature' vs. 'of neurosis' or 'studies' vs. 'is studied'*) and diacritized output. The popup window is the tagging of the verb *wa_yadorusu* "and _learn". Source: https://camel.abudhabi.nyu.edu/madamira/?locale=en

Fig 2.4 is a screenshot of a 14-word sentence tested with MADAMIRA. Tables 2.4.b and 2.4.b-bis detail the tagging of this sentence and the output for 5 of its explicit vowels (underlined); vowels are **bold underlined** if explicit in the input, but removed and wrongly recomputed by MADAMIRA; they are and **bold** if omitted in the input and wrongly computed by MADAMIRA. The grey-background columns display MADAMIRA outputs.

Table 2.4.b MADAMIRA vowelization and tagging output details for sentence in Fig. 2.4

| Line | Transliteration | Input Text | MADAMIRA Output: Diacriticized Text | (should be) | MADAMIRA Output: Diacriticized Text | Meaning | Meaning selected by MADAMIRA |
|---|---|---|---|---|---|---|---|
| 1 | Alfrq | الفرق | الفُرَقُ | Alfaroqu | Alfiraqu | the_difference | **the_groups** |
| 2 | AlJy | الذي | الَّذِي | AlGaJiy | AlGaJiy | that(masc-sing) | that (masc-sing) |
| 3 | t**u**Hdvh | تحدثه | تَحْدُثُهُ | t**u**Hoduvahu | t**a**Hoduvuhu | (she)makes-happen_it | **happens_it** |
| 4 | Hrkp | حركة | حَرَكَةٌ | HarakapN | HarakapN | (a) vowel | (a) vowel, motion |
| 5 | sgyrp | صغيرة | صَغِيرَةٌ | sagiyrapN | sagiyrapN | small | small |
| 6 | fy | في | فِي | fiy | fiy | in | in |
| 7 | mEnY | معنى | مَعْنَى | maEonaY | maEonaY | meaning | meaning |
| 8 | klimAt | كلمات | كَلِمَاتٌ | kalimAt**K** | kalimAt**N** | words (nominative) | of words (genitive) |
| 9 | mvl | مثل: | مِثْل | mivola | mivola | like | like |
| 10 | EiSAb | عصاب | عِصَابٌ | EiSAb**K** | EiSAb**N** | ligature | of ligature |
| 11 | wE**u**SAb | وعصاب | وَعِصَابٌ | waE**u**SAb**K** | waE**i**SAb**N** | and_neurosis | **and_of_ligature** |
| 12 | Ow | أو | أو | Oawo | Oawo | or | or |
| 13 | y**a**drs | يَدرس | يَدْرُسُ | y**a**dorusu | y**a**dorusu | studies | studies |
| 14 | wy**u**drs | ويُدرس | وَيَدْرُسُ | way**u**dorasu | way**a**dorasu | and_is_studied | **and_studies** |

Table 2.4.b-bis Complementary notes on MADAMIRA output. The line numbers refer to the lines of Table 2.4.b

| Line | Notes on the diacritics computed by MADAMIRA (wrong/correct) | Notes on agreement mismatch and other discrepancies |
|---|---|---|
| 1 | Selection of a wrong lemma *firaq* /*faroq* | *firaq*: broken plural of *firqap*. In this situation, words in grammatical agreement with this one are in the feminine singular |
| 2 |  | PRONOUN: agreement mismatch with the noun selected as coreferent (line 1): masc_sing/fem_sing |
| 3 | After the removal of *u*, selection of the wrong verbal lemma *Hdv/Ohdv*, "happen/makes_happen" | "happen" is an intransitive verb, the agglutination of a clitic pronoun (here, object pronoun) is wrong. |
| 8 | Wrong case ending *N* instead of *K* (nominative/genitive) |  |
| 9 | Wrong value of definiteness: 'construct state' (*mudaf*)[7], *mivola* | Mismatch between the features and the case-marking diacritic: if in the construct state, *mivol**a*** should be |

---

[7] The three values of definiteness in Arabic are definite, indefinite and construct state. A noun is in the construct state if it has an adjunct in the genitive.

| | is correct | in the genitive case *mivoli* |
| --- | --- | --- |
| 10 | *EisaAbN/EisaAbK* | case ending must be genitive instead of nominative |
| 11 | After the removal of *u*, selection the wrong lemma, although the other entry exists in BAMA | Case ending must be genitive (*-K*) instead of nominative (*-N*) |
| 14 | After the removal of *u*, selection of the wrong voice of the verb: active instead of passive | |

MADAMIRA removes all diacritics, recomputes them according to the BAMA lexicon and algorithm, and finally selects a solution from the available candidates: "*Input text enters the Preprocessor, which cleans the text and converts it to the Buckwalter representation used within MADAMIRA. The text is then passed to the Morphological Analysis component, which develops a list of all possible analyses (independent of context) for each word. The text and analyses are then passed to a Feature Modelling component, which applies SVM and language models to derive predictions for the word's morphological features*" (Section 3, Pasha et al., 2014).

In the example, four meanings (in **Bold** in Table 2.4.b) are wrongly selected by MADAMIRA. The agreement between the relative pronoun and the BP is incorrect (Table 2.4.b, line 2). The correct grammatical agreement between a broken plural and an adjective sets the adjective in the feminine singular. MADAMIRA finds correctly the related singular form, but systematically selects the masculine-singular form of an adjective following a broken plural instead of the feminine-singular form.

According to the authors, MADAMIRA has 86.3 % of words well diacritized, an improvement compared to 82.7%, which is the precision of Zitouni et al. (2006). On the other side, it has 84% of precision in disambiguation (EVALFULL). This means about two tagging errors per line in a text. In a pipeline of NLP, we estimate MADAMIRA useless with such an error rate.

To sum up, MADAMIRA computes erroneous vowels, omitted in the input; and it removes correct ones written in the input and replaces them by erroneous ones, which is more shocking since such errors are obviously evitable. Finally, its language model fails to capture some dependencies between adjacent words.

Like Madamira, Farasa (Mubarak and Darwish, 2014) removes first the presumably valid diacritics from the source text and recomputes autocorrected words according to its processing pipeline. It seems that the autocorrected words are recalculated based on "common typographical mistakes", such as the final *h/p* (Table 2.5, line 1) or y/Y (line 2), very likely combined with a rough frequency of tokens without taking into account word segmentation. In Table 2.5, we show three examples submitted to Farasa (http://qatsdemo.cloudapp.net/farasa/demo.html):

Table 2.5 FARASA: Three examples with *G* diacritics deletion and auto-correction

| Line | Input Text | FARASA autocorrected text | Transliteration | FARASA Transliteration | Meaning | FARASA Meaning selected |
| --- | --- | --- | --- | --- | --- | --- |
| 1 | سيّده | سيدة | sy**G**dh | sydp | master_his | (a) lady |
| 2 | التقيّ | التقى | Altqy**G** | AltqY | the_devot | (he) meets |

| | 3 | يحدّثونها | يحددنها | yhd**G**vwnhA | yhd<u>dn</u>hA | talk(they-masc)_her | Defines (they-fem)_her |
|---|---|---|---|---|---|---|---|

In the words in lines 1, 3, the reader must restore a gemination diacritic in `sy`**`G`**`dp` and `yhd`**`G`**`dnhA`, not explicitly given by Farasa resources; and in line 3, besides removing the valid *G* diacritic, the processing removed 2 other letters, replacing the masculine plural form by the feminine plural of another verb lemma.

Hamed et Torsten (2017) compare Farasa to Madamira: their Table 11 (annotated WER subcategories) shows that errors for both systems are **mainly** related to diacritics, 13/16 errors for Farasa and 14/18 for Madamira. The paper concludes: "*We find that FARASA is outperforming MADAMIRA in both evaluation modes, but that **in relaxed mode the simple dictionary lookup baseline is surprisingly strong**. In general, our error rates are much higher than the ones reported in the literature and we currently have no satisfying explanation for the difference*".

Zalmout & Habash (2017) present a model for Arabic morphological disambiguation based on Recurrent Neural Networks (RNN); "*adding learning features from a morphological analyzer to model the space of possible analyses provides additional improvement.*". Compared to MADAMIRA, the accuracy of the system with RNN improves from 85,6% to 90%. They evaluate the accuracy for out-of-vocabulary words separately, as 7,9%: globally, the accuracy is in fact 77%; therefore, the accuracy is almost 96% for words in the vocabulary. They conclude "*that enriching the input word embedding with additional morphological features increases the morphological tagging accuracy **drastically***". Nonetheless, a better coverage would increase even more the accuracy of the whole system.

"*When considering full analyses, we observe that our system still makes some errors in words where MADAMIRA is correct. However, the number of times our system is correct and MADAMIRA is not is over twice as the reverse (MADAMIRA is correct and our system is not)*". Explanations of why and how such dissimilarities and differences happen would be speculative. It seems the SVM approach of 2014 cannot benefit from the RNN approach in 2017, and reciprocally. ***This is a serious limitation for scientific improvements***.

## 2.5 Automatic diacritization with RNN (2015)

Abandah et al. (2015) present an Arabic diacritizer based on Recurrent Neural Network (RNN-LSTM). The processing is divided in two stages: the RNN transcribes the input into a fully diacritized sequence; then post-processing corrections are applied to overcome some transcription errors.

Since our purpose in this article is to propose linguistic resources with rich encoding that can be used in symbolic or statistical NLP pipelines, we describe below the related "light" linguistic operations in the post-processing stage.

The post-processing includes:
- *Sukun correction*: *o* (zero-vowel) diacritics are removed from the transcribed sequence. For example, the output *AlotGaAlibu* is corrected to *AltGaAlibu*[8].

---

[8] Abandah et al. (2015) does not respect the orthographic representation in his examples, so we have transcribed the examples given according to TB++ encoding which is a mapping one-to-one (cf. footnote 1).

- *Fatha correction*: The letter that precedes *A*, *Y*, *p* always has the short vowel *a* or *Ga*. If such a letter in the output sequence has a short vowel other than *a*, it is corrected to *a*. For example, the output *AltGuAlibu* is corrected to *AltGaAlibu.*
- *Dictionary-based correction*: "A dictionary is consulted to check whether the output word is in this dictionary. This dictionary is built from the training data and is indexed by the non-diacritized version of the word." The dictionary is 3 million words (or twelve thousand pages) – see Table 1, mainly from the "Tashkila collection of Islamic religious heritage Books". Such an index is rudimentary for diacritization, because of its low coverage.

Table 2.5. From Table 7 of Abandah et al. (2015)

|   | Target | Output | Notes target | Notes on Output |
|---|--------|--------|--------------|-----------------|
| 3 | yaSonaE-a | yaSonGaEa | Fabricates (he)-Subjunctive | Invalid word: invalid phonological sequence 'onG' |
| 5 | la tar-uwanGa haA | litarawonihaA | to_see-(you-mas-plu-Energetic) her | invalid token |
| 6 | wal**a**A | walA**a** | and_not | invalid typography: *A* is never with vowel, *a* must precede |

Table 2.5 shows 3 sample sequences that have errors, out of six in Table 7 of Abandah et al. (2015). We show that the use of linguistic resources allows for avoiding such errors:
- *yaSonGaEu* is an invalid token that may be detected if a dictionary offers the valid vowelized candidates to *ySnE*. Moreover, this word form **breaks** a major phonological rule: the diacritic *o* cannot precede a geminated consonant as in *onGa*.
- *li_tarawoni_haA* is ungrammatical[9] with an impossible verbal suffix –*awoni* instead of –*awona*. The vowelized output for *ltrwnhA*, لترونها should be *la_taruwanGa_haA*[10]. The imperfect in the energetic mode is a rare form in Arabic. Here, it is the inflected form of a frequent verb meaning "to see"; but the two agglutinations make this form even more rare in current corpora. This token occurs in the Koran, and we have found only one occurrence in the ArabicCorpus, occurring in a quotation of the same Koranic verse. However, our resources predict this rare agglutinated form.
- Finally, it outputs *wa_lAa* instead of *wa_laA*, which is a typographical error.

Abandah et al. (2015) is one of the very few experimentations that makes almost no use of Arabic linguistic knowledge. Such extreme usage of Machine Learning techniques in Arabic NLP shows bluntly its flaws and its limits. Statistical techniques are able to learn from aligned data made of character strings such as (*ySnE, yaSonaEa*), but they are unable to learn that *yaSonaEa* is a verb and its lemma is *SanaEa* with such data. It is no surprise that without comprehensive linguistic knowledge, such technology generates invalid word forms, even worse, it generates strings that are phonologically and typographically invalid. In addition, building a lexical resource is a better investment than dedicating an equivalent effort to manually annotating a corpus, because a comprehensive dictionary is valid for long and for many domains. The existing entries of the dictionary need not be edited as long as the behaviour of the words don't change, whereas a new corpus must be annotated every time you change domains.

---

[9] If the subordinate conjunction *li* is retained, *li_tar-awona_haA* is ungrammatical too, because of the presence of *na*.

[10] This token is validated by our resources (agglutination grammars and full-form dictionary): our parser restores the vowels, recognizes three agglutinated segments and relates the stem with the verbal lemma "to_see": لَتَرُوَنَّهَا {لَ.PART_la} {رأى.V:aI2mpE}تَرَوُنَّ {هَا.PRO+Ppers+Acc:3fs}. (see, Neme, 2011). For all these 3 examples, if our resources are applied upstream in an NLP pipeline, they provide the right candidates; if downstream, they reject the ungrammatical output forms.

Finally, Abandah et al. (2015) admit they *"expect that providing the **morphological analysis** of such words to the RNN (Recurrent Neural Networks) would provide it with better information to achieve higher accuracy"*.

## 2.6  AlKhalil-2 resources (2016)

Boudchiche et al. (2016) present AlKhalil-2, a second version of AlKhalil-1 (Boudlal et al., 2010), a morpho-syntactic analyser for words taken out of context. AlKhalil-2 recognizes successfully partially or fully vowelized forms and eliminates incompatible analyses. The output provides for each word: a lemma field (inexistent in AlKhalil-1), rich inflectional attributes, traditional derivational POS labels, and some semantic labels proper to traditional Arabic morphology[11], such as temporal-locative nouns, associated usually to some derivational patterns. Finally, output labels are wordy (and in Arabic), which hinders integration in a NLP pipeline, as compared to mnemonic abbreviations.

The lexicon is in XML format and based on a root-and-pattern approach similar to SARF (Al-Bawab et al., 1994). Like SARF, the AlKhalil-1 algorithm for identifying forms is based on root-and-pattern morpho-phonological rules that apply to all the entries of its lexicon; whereas AlKhalil-2 operates on the basis of a multi-stem approach similar to BAMA (proclitics-stem-enclitics). AlKhalil-2 is written in Java and evaluated on a vowelized corpus containing mainly Islamic religious heritage and old classical books, with a relatively small amount of diacritized Modern Arabic texts.

Compared to AlKhalil-1 (cf. Neme & Laporte, 2013, Section 2.4.3), AlKhalil-2 improved its lexical coverage and its speed also improved seriously to 632 word/second[12]. AlKhalil-2 is even quicker when analysing fully vowelized text since the text is less ambiguous.

AlKhalil-2 segments agglutinated morphemes correctly and associates generally accurate inflectional attributes to words. The singular form (lemma field) is associated to its broken plural (BP) form, which was not the case in AlKhalil-1. Some of the awkward surface patterns in AlKhalil-1, such as *FaALa* فال associated to قال *qaAla*, were standardized to *FaEaLa* to correspond to the traditional patterns, but many awkward others still remain. For some difficult cases, more accuracy and improvements are necessary in computing the associated pattern. For example, with some more difficult BP[13] forms involving two or more morpho-phonological alternations, the association of singular form fails, for example in *barobariyG* (singular), *baraAobirap* (BP) "barbar(s)".

---

[11] The derivational tradition that associates semantic features to patterns is not reliable. As Al-Khalil-2 takes for granted this traditional morphology, it inherits the same flaws: for instance, it labels *muxaTGaT*, "plan, plot" مخطط as a temporal-locative noun.

[12] AlKhalil-2 performance is calculated on the basis of word types in texts not word occurrences. Words in a text are sorted; then the sorted list of word types (agglutinated or not) are labelled and presented to the user. However, the standard in NLP is to associate to each word occurrence the adequate labels, to keep the pair occurrence/labels text order. The output presentation is not standard in NLP.

The lexicon contains 215,508 lemmas: 42,656 for verbs and 172,852 nouns. The lexicon contains two root files for verbs and nouns with 7,500 roots each. These root bases generate 2,197,962 stems related to nouns and 1,903,541 stems related to verbs. Even if the authors standardized the patterns in the result presentation, behind the scene the concept of "surface pattern" remains in Al-Khalil-2. The lexical database contains a `VoweledStemCanonicPatternVerb` file with 1,756 vowelized patterns for (surface) stems of verbs. The `VoweledStemCanonicPatternNoun` file contains 8,042 vowelized patterns for (surface) stems of nouns (Boudchiche et al., 2014, Tableau 1, Boudchiche et al., 2016). There are two files for clitics: proclitics (67 compound elements, see Boudlal, 2010; Section 4.2) and enclitics (68 elements), sub-categorized by POS for nouns, verbs and common to both, as in BAMA.

The procedure for lookup into the lexical resource is complex with more than 20 steps: removing the diacritic but keeping a copy for checking incompatibility; operating a segmentation based on clitic compatibilities; analysing the stem for each valid segmentation:

- scanning non-derived word first (proper nouns);
- then scanning the stem of nouns (in five steps);
- then the stem of verbs (in five steps);

excluding invalid analyses via clitic compatibilities; excluding other analyses by using typographical rules. The result restores for each word the vowelized surface form with a rich tagging including root, pattern, POS and feature values, presented as CSV or XML format.

AlKhalil-2 is a new version of the lexicon of SARF and our remarks (Neme, 2011) still apply to it: "*The SARF project (Al-Bawab et al., 1994, http://sourceforge.net/projects/sarf/) is based on root-and-pattern representation. Starting from three-and four-consonant roots, it can generate Arabic verbs, derivative nouns, and gerunds, and inflect them. . […]. The project uses conventional programming techniques with the Java language and roots encoded in XML files. […]. The patterns are hard-coded in the form of Java code. […]; in addition, updating and correcting the language resource included in source code is complex since it involves two expertise: an Arabic linguist and a programmer; updating data and updating source code obey to different professional practices.*"

Besides, the number of 'voweled stem canonic patterns' for verbs and nouns is nearly 10,000. One may wonder how so many "stem patterns" are obtained and managed, and if there is a consensus in the team (linguists and computer scientists) around the (automatic maybe) attribution of such a "meta-morpheme" to each surface form. Moreover, many auxiliary fields are added to AlKhalil-2 databases, which makes it more complex.

---

[13] The coordinator of AlKhalil-1, Mansour Al-Ghamdi asked Alexis Neme during a conference in Beirut to evaluate AlKhalil-1. In May 2012, Alexis sent him an evaluation report (4 pages of technical report with annotated output from Al-Khalil1 in an Excel sheet). In this report, Alexis formulated such critics: awkward patterns, absence of the lemma field, etc. It seems that such critics were partially taken into account in AlKhalil-2.

Boudchiche et al. (2016, Section 5) claims *"AlKhalil-2 analyzer achieves a speed close to that of the fastest analyzer (632 words per second against 685 for BAMA analyzer). However, the speed coverage ratio is largely in favor of Alkhalil2 analyzer"*. However, the difference in speed is rather due to the fact that the BAMA lookup algorithm is written in PERL, an interpreted language (rather slow); whereas AlKhalil-2 is written in Java, a compiled language.

In 2012, in order to compare our verbal lexicon, we tested Al-Khalil-1 on the first 553 occurrences of verbs of the same test collection extracted from the Nemlar corpus (Neme, 2011). 42 occurrences of verbs were unrecognized, which represents an error rate of 7,6 % in the lexical coverage of verbs. With Al-Khalil2, our evaluation noted a strong improvement in the verbal coverage with a fault rate down to 0.5%.

For global coverage, we evaluated Al-Khalil-2 lexical coverage with the same corpus (11,950 words) used for evaluating Arabic-Unitex (cf. 5.3.1). Before running the test, we changed all *I* to *A*. The coverage is less than 88% for Modern Standard Arabic texts. We repeated the experience with other MSA texts and found coverages ranging between 87% and 93%. Many common relational adjectives are missing such as "terrorist", "colonial" "Zionist"; singular forms are covered but not broken plural forms as common as "turtles" and "bishops". Moreover, although the University of Oujda is in Morocco, the words *Amazigh, Amazighian* are not in the lexicon.

## 2.7 Automatic diacritization with AlKhalil-2

Using AlKhalil-2, Chennoufi & Mazraoui (2016) present a diacritizer that uses *"a hybrid system for automatic diacritization of Arabic sentences combining linguistic rules and statistical treatments"*. The processing is divided in 4 stages:

- for each word, AlKhalil-2 outputs diacritized candidate form/tag pairs, out of context;
- phonological/syntactic rules are used to eliminate invalid surface diacritized forms and/or morpho-syntactic analyses of a word;
- HMM algorithms determine the most probable diacritized sentence;
- finally, the system deals with words not analysed by AlKhalil-2.

Examples of rules of step 2:

- Phonological rules: two *o* (zero-vowel) diacritics in two consecutive syllables are not allowed in Arabic, so that *mino (A)lokitaAbi* (from the book) becomes *mina (A)lokitaAbi*. This rule is in cross-word diacritization, where a word ends with *o* and the following word begins with the determiner *Al-*. Thus, this rule relies not only on phonology but on segmentation and tagging, as well.
- Syntactic rules: <PREP><NOUN:genitive>, meaning that after a preposition only the genitive case ending is allowed; for example, *mina Alomadorasati* (from the school) is a valid utterance while *mina Alomadorasata* is not valid. Similar rules are implemented for <CONJ-SUBORDINATION> <VERB>, …

The system also includes a typographical standardization[14] of diacritics (Section 4.2.1): "*The tanween fatha sign with the letter Alif " ا "/A/ has two forms of writing: one before the letter (سَلاًمَا salaAm**FA** (peace)) and the other after the letter سَلاَمًا salaAm**AF**). The second form has been adopted*" [15]. In addition, the point 1) in the same section includes 3 occurrences of *AlomAlyziywna* 'the-Malaysians', instead of the correct form *AlomAlyz**G**iywna*, missing the gemination mark *G*. Such repeated errors indicate carelessness for linguistic data. Nonetheless, this does not lessen the value of the experiments and evaluations of the HMM in diacritization with or without rules.

Table 2.7. Comparison between Arabic automatic diacritization systems[16] (Chennoufi, Mazroui, 2016, from Table 3, 4). WER1/2 = Word Error Rate with or without case ending diacritics

| System | WER1 | WER2 |
|---|---|---|
| 1st assessment | | |
| AlKhalil-2-HMM | 8.29 | 4.10 |
| AlKhalil-2-rules-HMM | 6.28 | 2.58 |
| 2nd assessment | | |
| MADAMIRA-SAMA-SVM | 27.29 | 16.14 |
| AlKhalil-2-rules-HMM | 6.22 | 2.53 |
| 3rd assessment | | |
| Abandah et al. (2015)-RNN (Tashkeela corpus[17]) | 5.82 | 3.54 |
| AlKhalil-2-rules-HMM (Tashkeela corpus) | 4.45 | 1.86 |

Each assessment in the Table 2.7 reproduces the same evaluation metrics. The first comparison is between AlKhalil-2-HMM with or without rules and shows a better result (+2%) with rules.

---

[14] In newspapers, the most frequent variant is *–AF*; literature magazines (such as http://al-adab.com/, Evaluation *Section 5.3.1*) and reference books adopt the normative variant *–FA*, since the variant *–AF* is considered by normative grammarians as erroneous. In this case, the choice of variant (or typography) depends on editorial practices in a printing industry.
[15] Default rules for diacritics in Al-Khalil-2 are similar to Neme (2011, section 4.2), implemented but documented in the Unitex User Manual.
[16] The paper includes also an evaluation of the MS-Office plug-in *Arabic Authoring services*, with word error rates (WER1 and WER2) of 20.56 and 11.18, better than MADAMIRA. We do not have access to the description of the *Arabic Authoring services*; nonetheless, the better performance of the plug-in is partly due to the lexical coverage of the Arabic resources of MS-Office, better than the embedded SAMA in MADAMIRA.
[17] http://sourceforge.net/projects/tashkeela/

About the comparison with MADAMIRA and Abandah et al. (2015), Chennoufi & Mazroui (2016) conclude that the good performances of the system are consequences of *"combining morphological analysis, syntactic and diacritic rules and* [of the] *large size of the corpus (used in statistical processing)"*.

## 2.8 Conclusions and perspectives

As Attia et al. (2011) underline, *"The quality and coverage of the lexical database determines the quality and coverage of the morphological analyser, and limitations in the lexicon will cascade through to higher levels of processing"*. This is true for diacritics too. The accusative suffix *-F* (pronounced [an]) is likely to help in the disambiguation of words, the gemination diacritic in selecting the right lemma of a verb (causative, for instance) or a noun, and the presence of a *u* after the first root consonant in the detection of a passive. Such inconspicuous information is valuable for disambiguation.

AlKhalil-2 eliminates analyses incompatible with the partially vowelized word but through lookups in several XML databases. Chennoufi & Mazroui (2016) demonstrate that *"combining morphological analysis, syntactic and diacritic rules used in a pipeline with statistical processing produces better performance than other systems",* including the RNN approach. No matter the approach, symbolic or statistical, one may expect a better result in disambiguization or vowelization with a better lexical resource in an Arabic NLP pipeline.

Hamdi (2012) demonstrates that statistical approaches were unable to give a satisfactory solution for partially vowelized words, whereas symbolic approaches propose a solution with disarming simplicity.

Our solution, which was implemented in November 2010, is similar to Hamdi's (2012). Nonetheless, Arabic-Unitex was built on a more radical basis: from the beginning, the lookup procedure ***retains only the candidates compatible*** with a partially diacritized word. The procedure uses a compressed finite-state automaton (FSA) and accesses the fully vowelized resource to discard the paths incompatible with the diacritics present in the text.

Arabic-Unitex uses FSTs intensively for inflection and takes into account all morphological and orthographical alternations to achieve a large lexical coverage of Arabic. The lexicon has been built and encoded manually. Arabic-Unitex consists of 76,000 lemmas and is inflected into 6 million fully vowelized forms, which are stored in an FSA data structure for fast retrieval through a lookup procedure. We evaluate the potential of recognizable agglutinated forms to more than 500 million valid forms if we count only fully vowelized forms, and to several billions of recognizable and valid partially vowelized forms.

In what follows, we will present briefly the overall architecture of Arabic-Unitex.

## 3 General presentation of Arabic-Unitex

Arabic-Unitex is a lemma-based, fully vowelized language resource with straightforward inflectional encoding based on the Semitic grammatical tradition and extended by independent agglutination grammars. In 2010, being aware of the four complications (cf. Section 1) facing the Arabic computational morphology, we adapted Unitex programs and tools to Arabic traditional representation so that the resources may be more easily read and maintained by Arabic linguists. We have adjusted Unitex programs to deal with:
- inflection with Semitic patterns or infixes;
- agglutination of proclitics/enclitics;
- partial vowelization.

## 3.1 The PRIM Model

Inspired by the Semitic traditional root-and-pattern model, our model for Arabic morphology requires detailed lexical representation as well, but uses at the same time up-to-date algorithmic techniques (FSTs). Neme & Laporte (2013) introduce the ***pattern-and-root inflectional model*** (PRIM) for Arabic morphology. We define a *pattern* as a template of characters surrounding the slots (place-holders) for the *root* letters. Around the slots, patterns contain short vowels, and sometimes consonants or long vowels.

The breakthrough lies in the reversal of the traditional root-and-pattern Semitic model into pattern-and-root, giving precedence to patterns over roots. Traditionally, the analysis of an Arabic word begins by assigning it an **etymological** root, and the rest is the pattern[18]. We begin by instead recognizing the **inflectional** pattern of the word, and the remainder is the root. In the traditional analysis, the pattern combines derivational and inflectional information, including all the derivation of the word from its remotest root. With our innovation, it is purely inflectional. This change keeps the expressiveness of the traditional model, which has been tested and validated during ten centuries; additionally, it enables faster identification of the verbal entry, its root and its pattern, with a smaller margin of error; moreover, it avoids the definition of several hundred interdependent morphological, phonological and orthographic rules.

Pattern-and-root inflectional morphology is adequate to Arabic morphology. We keep inflection apart from derivational morphology. The PRIM inflectional sub-taxonomies for verbs, suffixed plural and BP are simple, methodical and detailed; they avoid shortcuts or over-simplifications. The PRIM model complies with the conventions of the Semitic traditional morphology and is understood quickly by Arabic-speaking linguists. The lexicon is organized in fully vowelized lexical entries, like traditional dictionaries; and not in stem entries, as in the multi-stem approach. A lexical entry in traditional dictionaries is a lemmatized entry as well, but entries with the same etymological root are indexed under this root, and roots are ordered alphabetically.

In the PRIM model, a pattern is a simple sequence of consonant slots, consonants and vowels (short or long), but is not used to represent a meaning or morpho-syntactic features attached to patterns. In PRIM, a root is merely a sequence of letters (usually consonants). Orthographical variations of the glottal stop are encoded in the same way. Root letter substitutions and insertions are restricted to *w*, *y*, *A*, and to glottal stop allographs. We deal with morpho-phonological alternations in a factual way: inflected forms are generated from their observable surface lemma, and not from a "deep" or "underlying" root.

---

[18] Smrz (2007) converges with us on the definition of root and diverges on the definition of pattern: "*The 'root' should not be understood in the sense of Semitic linguistics. Rather, it is the core lexical information associated with the lexeme and available to the inflectional rules.*" (p.31). Smrz creates the concept of morphophonemic pattern (surface pattern) which creates numerous patterns awkward to native speakers: "*Morphophonemic patterns and their significance for the simplification of the model of morphological alternations*" (p.13).

An inflectional transducer is associated with each inflectional class in the taxonomy, and it generates all the inflected vowelized forms of any lemma in the class. Each lexical tag is accurate and informative and its format consists of a lemma followed by a set of feature-value pairs. Agglutinated clitics are analysed without the generation of artificial ambiguity. Clitic-agglutination grammars are described independently from inflection, in separate grammars. Morphological analysis of Arabic text is performed directly with a dictionary of words and without morphological rules: all orthographical variants are registered in the dictionary, which simplifies and speeds up the process.

The main challenge was to elaborate the inflectional model of pattern-and-root morphology based on Semitic grammatical tradition and our critical reading of Beesley's work (1991-2001), a generativist forerunner in Arabic computational morphology. If one can find attempts to build a systematic taxonomy for verbs in the Arabic morphological tradition already in the 10[th] century, it is the first time that the broken plural gets a straightforward and elegant representation based on three new principles crafted for encoding Semitic morphology. Moreover, they were complemented by concatenative encoding for regular suffixation to depict all aspects of morphological representation.

## 3.2  A full-form inflected dictionary

A line encodes one lexical entry in our lemmatized lexicon. The encoding contains a lemma followed by grammatical codes, and optionally comments. In order to facilitate direct human reading of the entry, the lemma is separated from the code by a simple comma, and the code from the comments by a slash. For regular plural, also known as sound plural, the inflectional transducer is designed to be used by the generator of inflected forms in the concatenative mode, which is the default mode.

The grammatical code contains sub-fields for singular, gender and plural, separated by hyphens:
```
nufaAyap,N00ap-f-At/   نفاية       'rubbish'
```
/ singular ending in *-ap* ("teh marbutah" in Arabic); feminine; plural suffix in *-At*
```
manaAx,N0000-m-At/    مَنَاخ        'climate'
```
/ singular with no particular suffix; masculine; plural suffix in *-At*

Our lemmatized lexicon produces fully vowelized forms by using FSTs based on a Semitic-style taxonomy for verbs (Neme, 2011) and nouns (Neme & Laporte, 2013).

The output format of an FST is `surface-form,lemma.V:feature-values` as in:
```
takotubu,ktb.V:aI3fsN /active-Imperfect-3rd_Pers-fem-sing-iNdicative
```
The '/' character comments out the text that follows it up to the end of the line.
For verbs, the *feature values* are detailed as in traditional morphology and in the following order:
- Voice: active (a), passive (b);
- Tense: Perfect, Imperfect, Imperative (Y);
- Person: 1, 2, 3;
- Gender: masculine, feminine;
- Number: singular, dual, plural;
- Mode: indicative (N), Subjunctive, Jussive, Energetic.

For nouns and adjectives, the *feature values* are in the following order:
- Gender: masculine, feminine.
- Number: singular, dual (d), sound plural (p), broken plural (q).
- Definiteness: definite (D), indefinite (i), and construct state (a).
- Case: Nominative, Accusative, Genitive.

The order between features is not significant, but our resources respect a fixed order, in order to facilitate human reading and therefore checking.

'*Distinct codes are required for broken plural (q) and suffixed plural (p) because rules of agreement between a plural noun and an adjective, a participle or a verb depend on whether the noun is a BP or a suffixed plural (Neme & Laporte 2013, pages 243-245).*'

### 3.3 Delimited Word Forms (DWF) grammars

A word delimited by spaces or punctuation symbols (DWF) is composed of a sequence of segments. A word or DWF is described in our resource of Arabic as the undelimited concatenation of clitics around an inflected form. Agglutination of morphemes in a word is represented by grammars. Each segment in a word will be called a morpheme[19]. The combination of a sequence of morphemes obeys a number of constraints which are expressed by a POS agglutination grammar. For instance, a verbal word is composed by one morpheme *<V>* or the concatenation of up to 4 morphemes as in:

    *<CONJC> <CONJS> <V:inflected> <PRO+accusative>*

where <CONJC> is a coordinating conjunction, <CONJS> is a subordinating conjunction and *<PRO+accusative>* an agglutinated object pronoun.

*<CONJC>* combines freely with any inflected verb. The *<CONJS>* constraints the verb to the imperfect subjunctive or to the jussive. Finally, an inflected verb is often insensitive to the agglutinated pronoun (i.e. its form is not affected) but some forms are sensitive: for example, forms with a glottal stop as the third root consonant (for verbs, see Neme, 2011, Section 4.1; for nouns, see Neme & Laporte, 2013, Section 8).

In BAMA, agglutination of verbs is formalized by the following:
```
[<CONJC>][<CONJS>]<inflexional-prefix><V-stem><inflexional-suffix>[<PRO+accusative>]
```

where <V-stem> is the string common to a subset of inflected forms vis-à-vis the concatenative operations and where the morphemes between [] are optional.

Both Arabic-Unitex and BAMA provide a segmented and tagged morphemic representation of a text. However, there are 2 essential differences: (1) Arabic-Unitex segmentation is closer to tradition and (2) Arabic-Unitex lemma grouping is closer to intuition: for example, singular and broken plural are grouped under the singular canonical form in Unitex, but under two stems (at least) in BAMA. With a better grouping of lemmas, lemma counts in a text are closer to the distribution of meanings. Therefore, we obtain a better representation of a document for applications such as automatic summarization and topic extraction.

---

[19]The morphemic status of some segments is controversial. The pattern, the lemma, the case ending may also be analysed as morphemes or morphs (find a detailed discussion in Smrz, 2007, morph *versus* morpheme). However, calling each segment a morpheme simplifies the description.

## 3.4 Building the dictionary based on a paradigmatic and taxonomic approach

In elementary and middle schools of Arabic-speaking countries, children are supposed to know by heart tables of conjugation and to compute all variations of a noun according to gender, number, definiteness and case. Irregularities are learned at school and related with two characteristics of the lemma: its pattern and the nature of its root consonants; then, once pupils have identified the lemma and the 'weak' root consonants (*A, w, y* and glottal stop), they learn to handle case endings, letter deletion, etc. according to syntactic context or the presence of an agglutinated pronoun. In addition, rules belong to a hierarchy of priority, but the hierarchy adopted by grammar textbooks is not always explicit, and sometimes fuzzy or messy. In our approach to computational morphology, the ordered and hierarchical rules learned at school were replaced by a formalized, operational grammar and a straightforward taxonomy. Each inflexional class in our taxonomy is provided with all the corresponding paradigmatic variations of forms, similar to the conjugation tables learned at school by children[20].

In our computational representation and tools, we have respected most of those habits and teaching methods, because they are widely shared by Arabic native speakers, and consequently by most potential descriptors of Arabic. For example, our citation form or lemmatized entry is similar to traditional dictionaries: the perfect 3rd person masculine singular for a verb, and the masculine or feminine singular for a noun or an adjective; and the description of inflection is similar to the traditional one.

We have adjusted Unitex tools in order to facilitate the encoding of paradigmatic variations. We have created two Semitic sub-taxonomies relative to verb variations and broken plural variations; each was split in two large sub-taxonomies related to the number of root letters: triliteral or quadriliteral, which is compatible with the traditional morphology. At the end, we have designed more than 1,150 inflectional classes; those for verbs and broken plurals are based on the pattern-and-root model, and those for suffix inflexion of noun and adjectives on the concatenative model.

As inflectional classes are numerous, the main challenge in our approach was to guess and assign the right pattern-class and root-subclass to each lexical entry when manually building or updating the dictionary. In order to facilitate this task, we designed the scheme to be straightforward and systematic, so that, for a given entry, linguists guess the associated class quickly. The sub-taxonomies are defined according to POS first, then to pattern classes and root subclasses:

- A straightforward verbal taxonomy for conjugation models with 460 classes (Neme, 2011).

- A straightforward broken plural taxonomy with 400 classes[21] for nouns and 50 classes for adjectives.

- The 250 remaining classes are dedicated to nouns and adjectives with suffixed plural and other POS classes. This number is comparable to the number of classes for French

---

[20] See also http://babelarab.univ-mlv.fr, site in Arabic, for displaying tables of conjugation of 15 400 verbs including a table with an agglutinated pronoun, two tables for active and passive participles, and an Arabic spell checker with a unique feature for detecting invalid/misplaced diacritics.

[21] Neme and Laporte (2013) inventoried 300 inflexional classes for BP; this inventory increased with the lexicon extension to 4200 lemmas with BP instead of 3200 in 2013.

resources in Unitex.

The manual effort[22] towards the building of the lexicon may be schematically split into the following tasks:

- Typing-in the list of lemmas based on reference lists and dictionaries (checked mainly in Abdel Nour, 2006, as a reference dictionary).

- Encoding each lexical entry: POS and inflectional class.

- Hand crafting the 1,150 main graphs representing the inflexional classes and correcting each of them by checking the generated output, in part manually and in part automatically.

- Adding active and passive participles to the 460 graphs of the verbal inflection: 54 forms for active and 54 for passive.

- Generating automatically regular deverbal nouns (almost 10,000) and the related relative adjectives (almost 10,000) based on verbal lemma (V61-V70, V41-V42), taking into account 'weak' root consonant (*A, w, y* and glottal stop) alternations. These lists were filtered semi-automatically and checked manually.

- Validating codes, correcting typo errors, adding more classes….

- Enhancing the lexical coverage by processing corpora and by encoding valid words not found by Unitex.

## 3.5  Enhancing Lexical coverage

Fig 3.4 exemplifies the work involved to deal with a neologism: داعشيّ, the denomination of ISIS members in Arabic, in order to illustrate the task of extending the lexical coverage. This lemma has millions of hits in Google search with its masculine, feminine and broken plural forms[23]: `daAoEiMiyG:ms`, `daAoEiMiyGap:fs`, `dawaAEiM:q` (broken plural), `daAoEiMiyGaAt:fp`. An inflexional class for this neologism does not exist in our lexicon; however, we have found similar classes for (a) a triliteral noun ending in *–yG* 'kurodiyG,$**N3yy**-g-FvEvL-OaFoEaaL-123/ kurd*', admitting gender inflection, and for (b) triliteral nouns with the same pattern for broken plural. We made an inflectional transducer for (c) by combining parts of (b) for the masculine plural, and parts of (a) for the rest of the paradigm (Fig. 3.4). We named the new transducer and class with a similar combination.

```
a) kurodiyG,$N3yy-g-FvEvL-OaFoEaaL-123/ kurd          أكْراد   كُرْديّ   كُرْديّ
b) taAobiE,$N300-g-FvvEvL-FaEaaLiB-1w23/ dependent    تَوابِع  تَابِع   تَأبِع
c) daAoEiMiyG,$N3yy-g-FvvEvL-FaEaaLiB-1w23/           داعشيَات  دواعش  داعشيّ
```

---

[22] The manual effort cannot be quantified with precision in man-years; however it was a part time (with ups and downs) occupation of 1 person from 2010 to 2016.

[23] Note that the suffixed sound plural form داعشيّون, *dAEMiyG-uwn* (33 500 hits, Google search in May 2018) looks somehow awkward to native speakers as compared to the broken plural (2 930 000 hits). BP is preferred for most new nouns and suffixed plural for most new adjectives (Neme & Laporte, 2013). Note also the BP diptotic case ending, Fig. 3.4 "`N:sfx:uaiuaa`", where the nunation is not allowed for indefinite; and the genitive case is with *–a* ending.

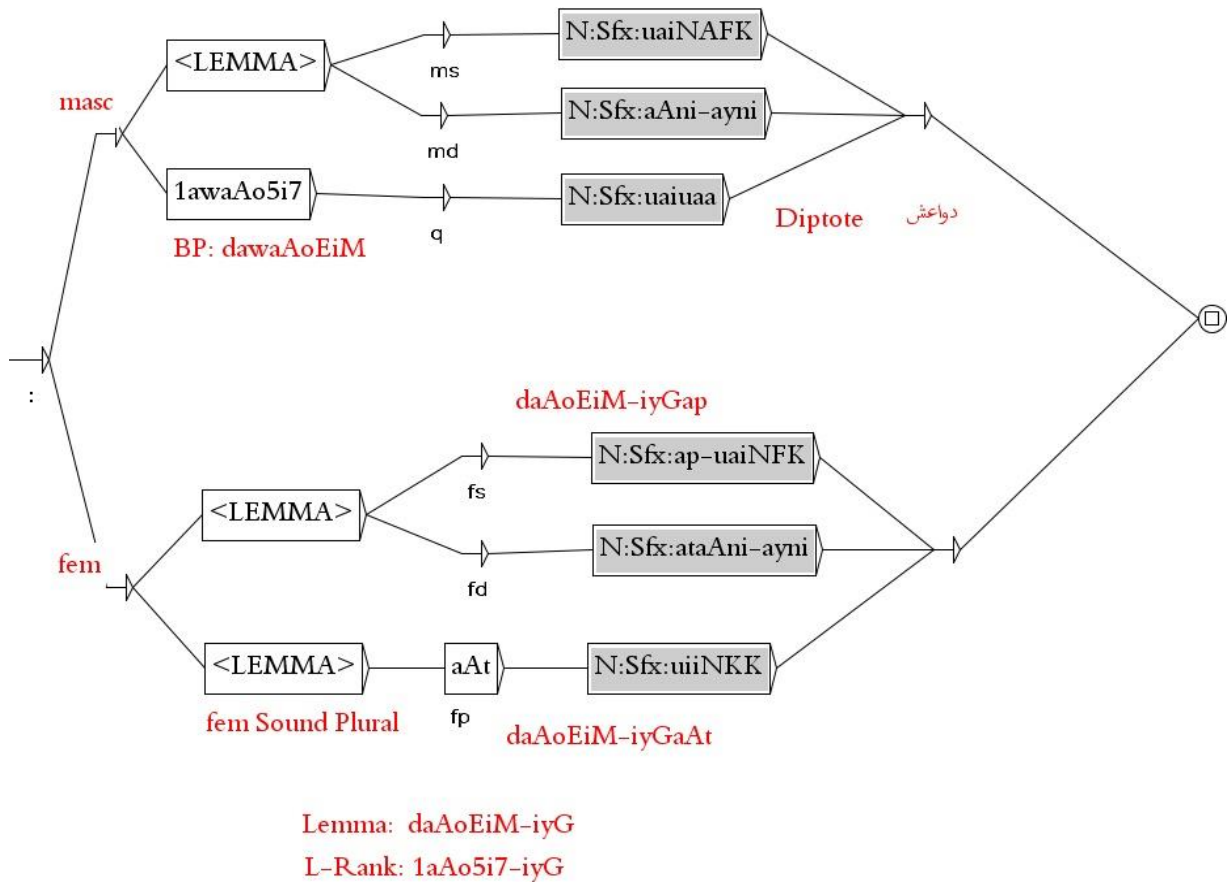

Fig.3.4. A new inflectional class for `daAoEiMiyG,$N3yy-g-FvvEvL-FaEaaLiB-1w23`

Even if many inflectional classes are replications with minor changes, creating 1,150 inflexional graphs (and 4000 sub-graphs, mainly for tenses and suffixed paradigms) was time consuming; besides, we have checked one by one the outputs of each inflexional graph. Summing up, the manual effort towards the building of the lexicon was to collect and type in each lemma, based on existing references dictionaries, verb lists, and results of corpus processing.

4    Vowel and vowel omission in Arabic-Unitex

4.1    Rules of vowel omission

Words in Arabic are often unvowelized and our system relies on our full-form inflected lexicon and agglutination grammars to restore the missing vowels. When Unitex uses a compressed Arabic lexicon that includes vowels, it is able to deal with unvowelized and with partially or fully vowelized words. If a word includes one or many diacritics, the lookup procedure extracts from the dictionary only the string candidates with the same diacritic(s) at the same position(s) as in the word, taking into account at the same time the predefined rules of diacritic omission.

A set of rules specifies in which conditions the lookup procedure tolerates vowel omission. In the Unitex folder for Arabic, the configuration file *Arabic-typo-rules.txt* defines rules for diacritic omission and other typography-related rules. The data distributed with Unitex contains this file with predefined rules suitable for usual printed text (see appendix); you can enable or disable each rule to cope with more restrictive or less restrictive standards. The predefined rules are designed to be used with a fully vowelized dictionary. The analysis restores the corresponding form(s) stored in the dictionary.

Each rule has the form RULE=YES/NO. Here are examples of rules:

- Rules of omission of one vowel/diacritic:
```
                                  / <dictionary_form> => <allowed_form>
                                  / <E> stands for the empty string
  fatha omission=YES              / a  =>  <E>
  dammatan omission at end=YES    / N  =>  <E> (N is pronounced [un])
                         / the kasra rule below is not in
                         /the predefined rules in the distributed data
  kasra omission=NO               / i => <E> rule disallowed
```

With the rules above, if *kitaAbN* is in the dictionary, *kitaAbN* matches it; *kitAb* and *kitaAb* also do; but *\*ktaAbN* doesn't, because *i* may not be omitted. [24]

- Rules of omission of two diacritics: When the word is fully vowelized, *G* is always followed by a short vowel (including *o* or a nunation). The following rules allow omitting *G*, but only if the vowel just after it is omitted too. Rules of Arabic script forbid to omit a *G* and write the vowel just after it:

```
  shadda fatha omission=YES       /   Ga => <E>
```
      *kat**Ga**ba => katba*
```
  shadda dammatan omission at end=YES   /   GN  => <E>
```
      *ruwsiy**GN**=> ruwsiy*       / رُوسِيٌّ

- Accusative marker inversion at the end of a word (*F* is pronounced [an]):
```
  fathatan alef equiv alef fathatan=YES /at the end -FA => -AF
```
      *kitab**FA** => kitab**AF***

```
  fathatan alef maqsura equiv alef maqsura fathatan=YES
```
      *fata**YF**  => fata**FY***         /FY =>YF

- Substitution of initial *O* or *I* (alif hamza) by *A* (bare alif):
```
     alef hamza above O to A =YES     / O => A
```
      ***O**akala => **A**akala*
```
  alef hamza below I to A=YES         / I => A
```
      ***I**ikotub => **A**ikotub*

- Rare diacritics:
  The presence or omission of the *R* superscripted variant of *alif* is handled by Unitex as well, e.g. in demonstrative pronouns.
```
    superscript alef omission=YES   / R => <E>, R superscript alif
```
      *h**R**JaA   => hJaA*       / هذا
      *AllG**R**hu  => AllGh*       /الله

---

[24] An asterisk '\*' indicates that a form is not in use in standard modern Arabic.

- Solar assimilation of *Al*: the assimilation of *l* to a coronal consonant (15 consonants/30) may be marked through the insertion of *G* after *Al<coronal-consonant>*: [25]

```
solar assimilation=YES
                                    /taAniy is in the dictionary
       AltaAniy                     / allowed, assimilation not graphically marked
       AltGaAniy                    / allowed too, assimilation graphically marked
```

The coronal consonants, which admit assimilation, are the following:

ت, t; ث, v; ج, j; د, d; ذ, J;      ر, r; ز, z; س, s; ش, M;
ص, S; ض, D; ط, T;                  ظ, Z; ل, l; ن, n; ه, h;

- Non-assimilation of *Al*: the assimilation of *l* to a non-coronal consonant (15 consonants/30) is disallowed in *Al<non-coronal-consonant>*:

```
lunar assimilation=NO   /check disallowed lunar consonant assimilation
                        /qamaru is in the dictionary
       Alqamaru         / allowed,
       AlqGamaru        / NOT an allowed form
```

The non-coronal consonants do not admit assimilation and are the following:

ء, c; آ, C; أ, O; ؤ, W; إ, I; ئ, e;  (all glottal stop variants)
ب, B; ح, H; خ, x; Z;  ع, E; غ, g; ف, f;
ق, q; ك, k; م, m;     و, w; ي, y; ا, A;

Table 4.1 illustrates the operation of the predefined Arabic typographical rules by giving the output of Unitex restoration. Each line in this table presents only one analysis, but in lines 3 and 4 Unitex produces several analyses.

Table 4.1. Restoration of vowels with the predefined rules. The TB++ and AR columns show the input

|   | TB++ | | AR | UNITEX Output | | |
|---|------|-------|-------|---------------|---|---|
| 1 | **Input** | **Notes** | Input | **Word with restored vowels** | **Lemma** | **POS:feats** |
| 2 | k**a**t**a**ba | All diacritics scripted | كَتَبَ | k**a**t**a**ba | كَتَبَ ｜ كتب | V:aP3ms |
| 3 | k**a**tb | 2 diacritics omitted | كَتب | kat**a**ba | كَتَبَ ｜ كتب | V:aP3ms |
| 4 | ktub | 2 omitted | كتُب | k**u**tub**a** | كُتُب ｜ كِتَأب | N:qaA |
| 5 | kt**i**b | 2 omitted | كتِب | k**u**tib**a** | كُتِبَ ｜ كتب | V:bP3ms |
| 6 | kat**G**b | 2 omitted | كتَب | katG**a**b**a** | كَتَبَ ｜ كتب | V:aP3ms |

---

[25] The letter *l* of the determiner is still written, but pronounced in the form of the following consonant.

| | | | | | | |
|---|---|---|---|---|---|---|
| 7 | kt**aG**b | Ga -> *aG | كتْب | | Unknown | |
| 8 | Al**qG**mru | wrong 'Al-' assimilation | القَمرُ | | Unknown | |
| 9 | Alqmru | no 'Al-' assimilation | القَمرُ | Alq**am**aru | قَمَر | N:msDN |
| 10 | Al**MG**msu | assimilation scripted | الشَمسُ | AlMGamosu | شَمْس | N:fsDN |
| 11 | **A**ErAbN | allowed variant of I (hamza-under-alif) | اعرابٌ | **Ii**EoraAbN | إغْرَابٌ | N:msiN |
| 12 | **O**ErAbN | wrong variant of I (hamza-under-alif) | أعرابٌ | | Unknown | |
| 13 | kitaAb**FA** | accusative marker, normative form | كتابًا | kitaAb**FA** | كِتَأْبًا | N:msiA |
| 14 | kitAb**AF** | allowed inversion | كتابأ | kitaAb**FA** | كِتَأْبًا | N:msiA |

Line 6 in Table 4.1 shows the form *katGb* where two vowels are omitted. Unitex dictionary lookup restores the vowelized full form *katGaba*, the related lemma *ktGb* and the morpho-syntactic tag V:aP3ms which means Verb in the active Perfect 3rd person masculine singular.

## 4.2 Inflected forms with short vowel variations

Arabic-Unitex takes into account short vowel variation in surface forms. This free variation affects the first vowel of some nouns. Three situations are common: *u/i/*a*, *a/u/*i* and *a/i/*u*; thus one may say *nufaAyap* or *nifaAyap* "rubbish" نفاية but not **nafaAyap*. The lexicon could record the two allowed vowelized forms in two lemmas, but we have chosen to encode this information in the inflectional transducers. This is less redundant and we avoid an artificial ambiguity between two lemmas in morphological annotations. Moreover, we also have the same allowed variations in the dual and in the plural: *nufaAyataAn/nifaAyataAn* "two pieces of rubbish" نفايتان; *nufaAyaAt/nifaAyaAt* نفايات "pieces of rubbish" for sound plural. The encoding of such variations was achieved for almost a hundred of lexical entries and needs to be completed.

In this section, we describe how we encoded lexical entries and inflectional transducers for nouns without vowel variant; then for nouns with vowel variant; finally, we present the special case of broken plurals and a similar variation observed in the suffixed plural of some feminine nouns.

### 4.2.1 Inflection without variant

The following three lexical entries undergo the short vowel variation in question, but here is an encoding that overlooks the vowel variation:

```
nufaAyap,N00ap-f-At/ نفاية          'rubbish'
manaAx,N0000-m-At/ مَناخ            'climate'
HaDaAnap,N00ap-f-At/ حَضانة         'kindergarten'
```

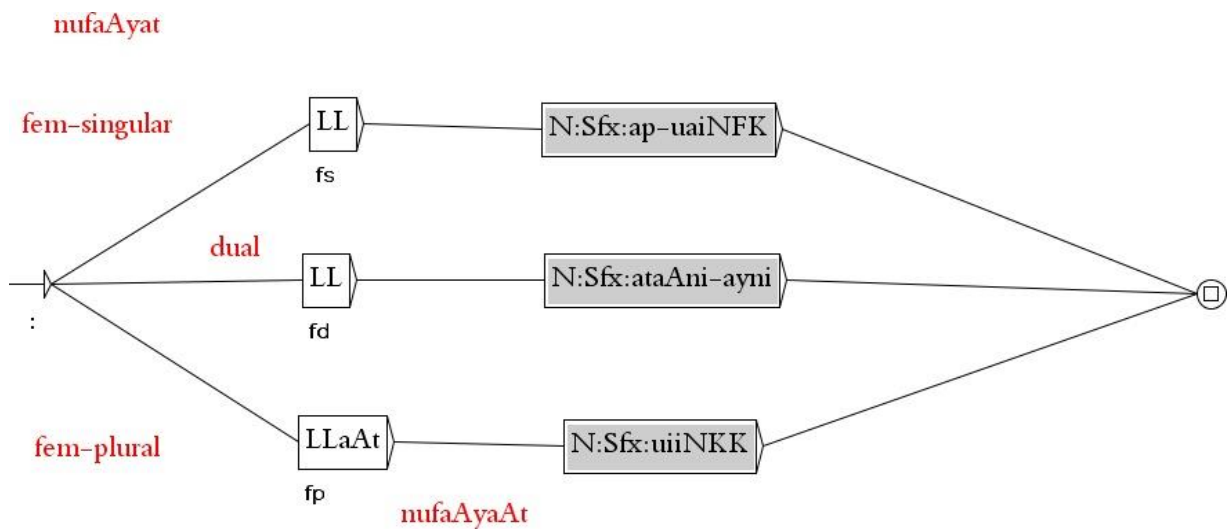

*Fig. 4.2.1.a. An inflectional transducer in the concatenative mode for nufaAyap*

Fig. 4.2.1.a shows the inflectional transducer for *nufaAyap* "rubbish"[26]. It contains three paths to produce singular, dual, and plural forms. The paths describe the suffixes to be added or removed to get an inflected form from a canonical form. The LL box (L is for Left shift) removes two letters from the end, here *ap*. The outputs (displayed under the boxes) are the inflectional codes to add to a dictionary entry[27]. A box not connected to another one is a comment or an explanation included in the transducer. A grey box is a call to a subgraph. In this graph, the subgraphs concatenate the suffixes of definiteness and case. For instance, the "N:Sfx:uiiNKK" subgraph (Fig. 4.2.1.b) represents the endings for the regular feminine plural.

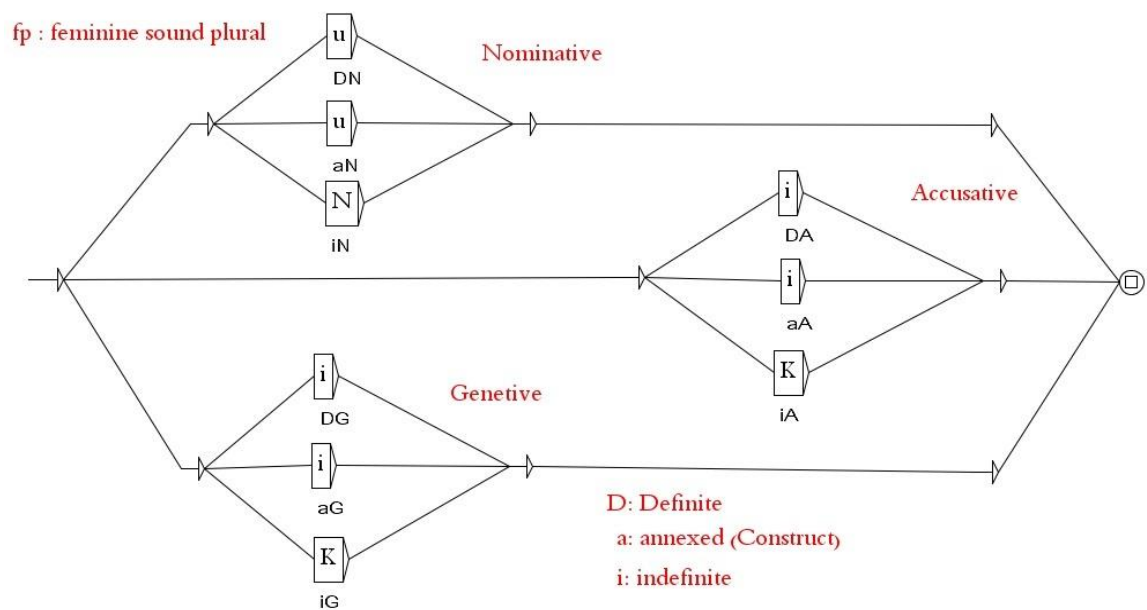

*Fig. 4.2.1.b. The N:Sfx:uiiNKK subgraph relative to the 9 variations of feminine plural*

### 4.2.2 Inflection with vowel variant

Here we describe our representation of short vowel variation. We use the generator of inflected forms in the Semitic mode, which is specified by the "$" symbol in the encodings below. We encode the vowel variation by inserting "_v_" in the grammatical code, where *v* indicates the alternate value of the first vowel. Below, the encoding of the same three entries as above, but with vowel variation.

```
nufaAyap,$N0_i_0ap-f-At/         نفاية          'rubbish'
manaAx,$N0_u_000-m-At/           مَناخ          'climate'
HaDaAnap,$N0_i_0ap-f-At/         حَضانة         'kindergarten'
```

---

[26] In this paper, we do not cover other free variations of short vowels such as the permutation of the vowels *a-i* in *minoTaqap* and *manoTiqap* "area" منطقة This variation may be written in the inflectional class as <1a2o3i4ap>.
[27] For a detailed description of inflectional transducers, see Unitex User Manual 3.1, Chap. 3.5, for concatenative and Semitic mode.

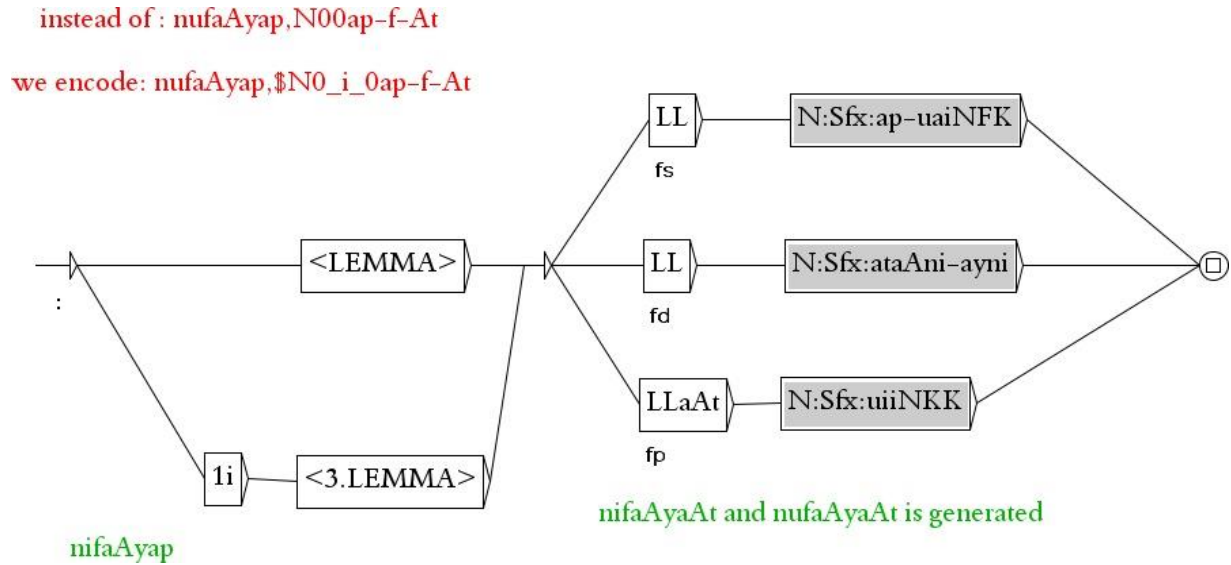

*Fig. 4.2.2.* An inflectional transducer in the Semitic mode for *nufaAyaAt/nifaAyaAt*

In the example (Fig.4.2.2), we have 6 paths: 3 paths inflect *nufaAyap* in the singular/dual/plural; they begin with the `<LEMMA>` operator, which retrieves *nufaAyap*, the lemma of the entry; the other 3 paths inflect *nifaAyap,* and they begin with the box 1i, which copies the first letter of the lemma, followed by the `<3.LEMMA>` operator, which copies the lemma from the third letter until the end. The `<n.LEMMA>` operator copies the lemma field from the $n^{th}$ position to the end of the field. The same three subgraphs representing suffixes are used in Figs. 4.2.1.a and 4.2.2, and in many other graphs.

The inflectional transducer produces both variants with *u* and with *i* as inflected forms of the lemma *nufaAyap* (in **bold** the example below). The inflectional transducer produces 54 inflected forms and associates them to the same lemma: 27 "standard" forms with *u*, plus 27 "variant" forms with *i*. The plural forms are the following output:

/standard with u                          / variant with i

```
nufaAyAatu,nufaAyap.N:fpDN        nifaAyAatu,nufaAyap.N:fpDN
nufaAyAatu,nufaAyap.N:fpaN        nifaAyAatu,nufaAyap.N:fpaN
nufaAyAatN,nufaAyap.N:fpiN        nifaAyAatN,nufaAyap.N:fpiN
nufaAyAati,nufaAyap.N:fpDA        nifaAyAati,nufaAyap.N:fpDA
nufaAyAati,nufaAyap.N:fpaA        nifaAyAati,nufaAyap.N:fpaA
nufaAyAatK,nufaAyap.N:fpiA        nifaAyAatK,nufaAyap.N:fpiA
```

The `<LEMMA>` operator copies the complete lemma field, no matter the number of letters in the field, and is useful for Arabic nouns and adjectives where masculine forms are generated by inserting vowels in the consonantal skeleton, whereas feminine forms are obtained by appending suffixes (Fig. 4.2.3.a). [28]

### 4.2.3 Vowel variant with broken plural

---

[28] These inflectional operators are useful also for an Austronesian language (cf. Unitex User manual Section 3.5.4 Inflection of Semitic languages): *In Tagalog, an Austronesian language that uses commonly infixes and reduplication for inflection, <LEMMA> and <n.LEMMA> may be used to produce verb tenses. The toy inflection grammar of Fig. 3.18 produces the perfect* kumain*, future* kakain *and imperfect* kumakain *of the verb* kain *"eat"*.

We have noticed this variation for the nouns *Euqodap/Eiqodap* "knot" عقدة, *gurofap/girofap* "room" غرفة, in the singular and dual, but also in the broken plural: *Euqad/Eiqad* "knots", *guraf/giraf* "rooms".

In the transducer for these entries (Fig.4.2.3.a), we use the <LEMMA> operator to copy the complete lemma field. The digits 1, 3, and 5 in the two boxes `1u3a5, 1i3a5` stand for the rank of the letter in the lemma in order to generate the broken plural (Neme & Laporte, 2013).

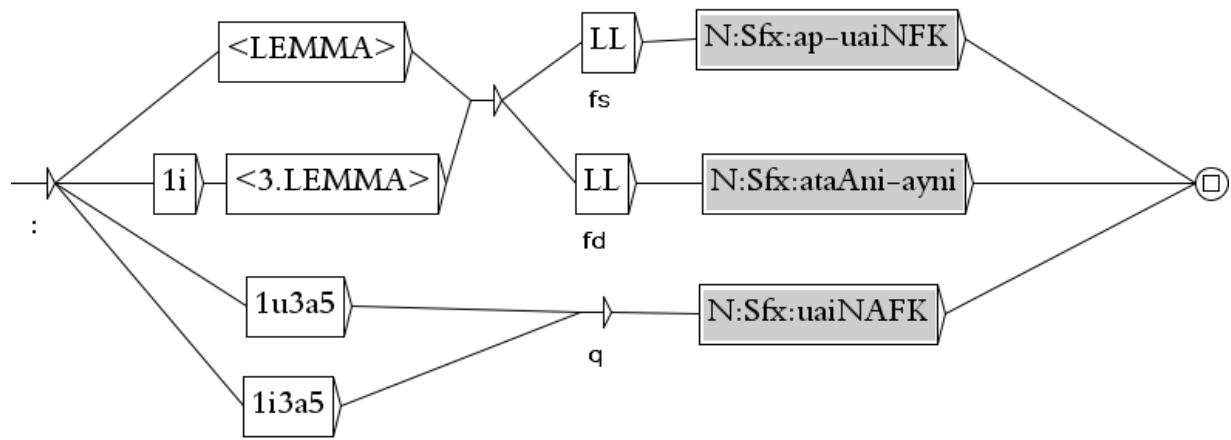

*Fig. 4.2.3.a. Inflectional transducer generating forms with vowel variation in the singular, dual and broken plural forms (in red, example in Arabic)*

Another case with a broken plural variant is *Saliyob* 'cross' صليب : we may say for the broken plural either *SilobaAn* or *SulobaAn* (Fig. 4.2.3.b) , but not *\*SalobaAn* صلبان . This pattern variation FuEolaan/FiEoLaan is frequent for broken plurals; still, not all nouns with the same pattern in the singular admit such variations: one may say *fusotaAn* "dress" but not *\*fisotaAn* or *\*fasotaAn* فستان.

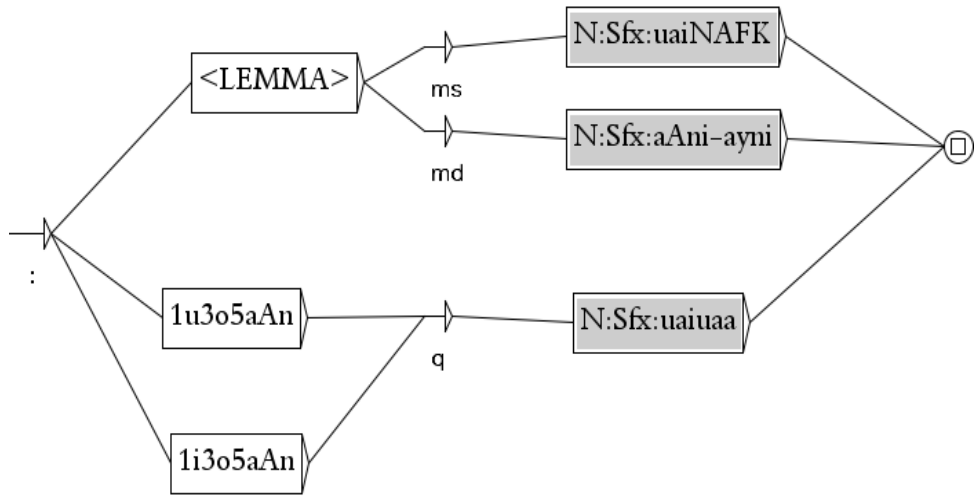

*Fig. 4.2.3.b. Inflectional transducer for broken plural variation for* Saliyob *'cross'* صليب, *we may say for the broken plural either* SulobaAn *(normative usage) or* SilobaAn

### 4.2.4 Suffixed feminine plural with *a/o*

Some feminine singular nouns such as *laSoqap* 'scotch tape' لصقة admit a variation in the plural (cf. Al-Ghalāyini, 2007, Vol 2, p.26): *laSaqaAt* vs. *laSoqaAt* لصقات (Fig. 4.2.4), or *Oazomap* 'crisis' أزمات , in the plural *OazamaAt* or *OazomaAt* أزمات . The sequence of operators LLLLaRaAt deletes from the end four letters, inserts *a*, copies a letter (here *q*) and adds *aAt* to produce *laSaqaAt* (L, R for Left, Right shift). Also note *suloTap/suluTaAt* 'authority' سُلُطَات, and more examples in Arabic in footnote [29] .

Instead of 27 forms, the transducer of Fig. 4.2.4 generates 54 surface forms (9x2 singular + 9x2 dual + 9x2 broken plural forms) and associates them to the same lemma.

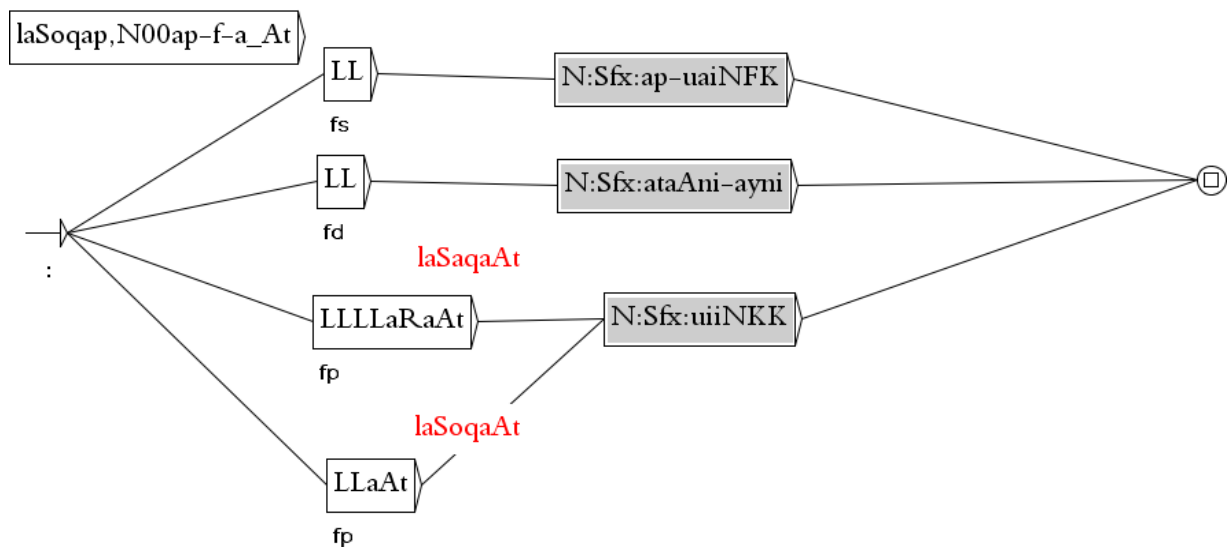

*Fig. 4.2.4. Inflectional transducer for variation of the plural with the suffix -aAt*

Tamazight, a Moroccan and Algerian language from the Hamito-Semitic family, has a similar phenomenon: the substitution of *e* (mute or pronounced schwa) by *a* before the plural suffix -*en* as in *izger/izgaren* "ox/oxes". This plural formation is called external plural[30] in this grammatical tradition.

## 5 Unitex - Arabic Lexicon

### 5.1 Tagset

---

[29] We identified many examples in our corpora: جُرُعَاتٍ خُطُوَاتٍ سُلْطَاتُ عُمَلَاتٍ نَدَبَاتٍ نَشَرَاتٍ نَشْرَاتٍ نَفَحَاتٍ هَجَمَاتٍ صَدَمَاتٍ صَفَحَاتٍ طَفَرَاتٍ حَلَقَاتٍ

[30] Nabil Chebieb, personal communication.

The following tables give an overview of the different codes used in the Arabic-Unitex dictionaries. These codes are meant to cover the morpho-syntax of Arabic simple inflected forms. For the open grammatical categories such as verbs, nouns and adjectives, all the inflectional values are detailed in appendix. They are consistent with traditional morphology, so that Arabic specialists can become quickly familiar with the tag set. The encoding is divided in three tables: POS (Table 5.1a), inflectional features (Table 5.1.b in appendix, with 360 combinations of inflectional features), and semantic-syntactic features (Table 5.1.c in appendix, with 30 syntactic and semantic features).

*Table 5.1a. Part Of Speech codes used in Arabic-Unitex*

| Code | POS in English | Encoded example | POS in Arabic | Arabic examples |
|---|---|---|---|---|
| <V> | Verb | <V:aI3msN> | فِعل | تتهمكون |
| <N> | Noun | <N:fsiG> | إسم | تُفَاحةٍ ، إمرأة |
| <NPr> | Proper noun | <NPr+Loc:fsDN> | إسم عَلم | دمشقُ |
| <A> | Adjective | <A:msiN> | صفة | صغيرٌ |
| <EL> | Elative, i.e. comparative and superlative | | أفعل التفضيل | |
| <ADV> | Adverb (indefinite accusative) | <ADV> or <V:FmsiA> | ظرف | واقفاً ، معاً |
| <PREP> | Preposition | <PREP+gen> | حرف (جر) | بَيْنَ |
| <PRO+Pdem> | Demonstrative pronoun | <PRO+Pdem:s> | اسم إشارة | هذان، هؤلاءِ، هناكَ، ذاكَ |
| <PRO+Prel> | Relative pronoun | <PRO+Prel-Hum:s> | ضمير | مِمَّا |
| <PRO+Pinterrog> | Interrogative pronoun | <PRO+Pinterrog+Hum:s> | إسم إستفهام | مَنْ ؟ |
| <CONJC> | Coordinating conjunction | | حرف عطف | أمَّا، أوْ، وَ، فَ |
| <CONJS> | Subordinating conjunction for verbs | | حروف النصب والجزم للفعل | لَنْ، لَمْ |
| <INTJ> | Interjection | | | واللهِ، |
| <DET> | Determiner Al- | | ال التعريف | الـــ |
| <INNA> | Governs accusative nouns | <INNA> | إنَّ وأخواتها | إنَّ أنَّ كأن |
| <PRTCL+Part_la> | Confirmation particle | | لام التوكيد | لَــــيَضربُ |
| <PRTCL+Part_sa> | Future particle before imperfect indicative | | سين المستقبل | ســـــيَضربُ |
| <PRTCL> | Any particle | <PRTCL+vocative> | أداة أو حرف | يا |

## 5.2 Size and parsing speed

The Arabic-Unitex lexicon of lemmas has been built and encoded manually and checked semi-manually. Its format consists of a simple line for each lemmatized lexical entry:

```
lemma,inflectional-code      / Notes
ktb,$V3au-123                / '$' indicates the Semitic mode
             / The encoding details are in Neme (2011)
kitaAob,$N300-m-FiEaaL-FuEuL-123
             / Broken plural    (See Neme & Laporte, 2013)
jamiyol,A0000-g-uwna
```

```
/A regular adjective admitting masculine and feminine inflection
/ with masculine plural in -uwna and feminine in -At
/ The inflectional transducer is in the concatenative mode
```

- The lexicon includes 76,000 lemmas and the full form language resource includes 6 million fully vowelized inflected forms.
- The lexicon has nearly 1,000 inflectional classes encoded in FSTs: 1,000 main graphs and 4,000 subgraphs
    - **15,400** verb lemmas
        - 4.1 million inflected forms including active and passive participles
            - including 550,000 inflected orthographic variants marked with +pro or +nopro for compatibility with enclitic pronouns
    - **41,500** noun lemmas including 4,200 with broken plural
        - 1.17 million inflected forms
            - including 125,000 inflected orthographic variants obligatorily with or without enclitic pronoun
    - **13,000** adjectives including 200 BP adjectives, and 200 elatives (such as "bigger")
        - 635,000 inflected forms
    - **6,000** proper nouns
        - 53,000 inflected forms (case and definiteness)
    - Several hundreds of entries with residual categories such as adverbs, pronouns, particles…
- For each POS, agglutination grammars are formalized in graphs restricting the combinatorics by using the inflectional attributes
- These resources potentially recognize at least **500 million** valid agglutinated words.

COUNTING PARTIALLY VOWELIZED FORMS

Equipped with our vowel-omission-tolerant lookup, the dictionary can store and identify a huge, theoretically infinite number of forms. Moreover, the presence of partially or fully vowelized words does not affect the speed of the analyser (section 6.2). In other words, <u>our data structure/algorithm is scalable</u>.

The lookup algorithm recognizes the form *yasotaqobilu*, for instance, and all partially vowelized variants with the omission of any number of vowels *yastqblu*, *ystqbil*, etc. and rejects as unknown incompatible forms such as *\*yas<u>a</u>taqobilu*, *\*y<u>i</u>sotqobl*.

We created a program to estimate the number of these potential partially vowelized forms by counting the occurrences of short vowels, *G* (gemination), *O* and *I* (*hamza* above and under *alif*) in each form in the inflected dictionary (6 million forms) and by computing the number of possibilities. Given that each vowel may appear or not, a fully vowelized form with 4 diacritics admits 16 possibilities of partial vowelization ($2^4$); a form with 5 vowels admits 32; and a form with 10 vowels admits 1024. The addition of such possibilities for the 6 million forms totals almost 250 billion partially vowelized forms. Moreover, if we include in the estimate the agglutination grammars (i.e. the agglutinated clitics which may have 1 to 4 vowels), this number can easily reach several trillion forms.

In addition, the system is able to discriminate between a huge set of correct forms and an even huger set of incorrect forms. The number of rejected forms is a theoretical, not an experimental, issue: in practice, the words that occur in real texts, either correct or incorrect, are much less numerous than the theoretical possibilities, either accepted or rejected. However, consider only the 4 short vowels *a, u, i, o*: one vowel is allowed at a given word position and the other 3 are incompatible with the fully vowelized form. The forms rejected by the algorithm for a word with 4 vowels are more than 81 ($3^4$); [31] with 5 vowels, they exceed 243; and with 10 vowels, they exceed 59 049 ($3^{10}$).

That is to say that ***an FSA is adapted to store and retrieve an infinity of string forms*** in a compressed file of about 10 Megabytes (see below about compression).

## 5.3 Evaluation

### 5.3.1 With a corpus with a high rate of vowelization

From *Al-adab* (http://al-adab.com/), a literature and critical essay magazine edited in Beirut since 1953, we have chosen three texts [32] (published in May, 2017, 60 pages): the first two are a political essay on democracy and an essay on the Syrian Civil War (2011-2017), written by Levantine writers from Lebanon and Syria, and representing together 15 pages; the remaining 45 pages are a discussion about Moroccan identity between six university professors and intellectuals from Morocco. Our choice of this corpus is motivated by the quality of its vocabulary, richer than in common newspaper texts, and the density of its authentic partial vowelization, which exceeds 33%, indicating a high level of editing process[33], achieved, we guess, by the writers, and controlled and enriched by the editor. This corpus allows us to test the Arabic-Unitex lexical resources and our lookup algorithm against partial vowelization that occurs spontaneously, independently from our lexical encoding. A carefully edited corpus with a high rate of vowelization provides a stricter evaluation than a corpus with a standard rate (3%), since each vowel written in the corpus is compared with vowels specified in the dictionary.

Our corpus is constituted of 11,950 words, 4,225 of them (*versus* 350 with a standard rate) with partial vowelization: 7,725 with no diacritics (64,6%), 3,886 with one diacritic (32.5%), 328 with two diacritics (2,74%) and 11 with three diacritics (0.1%). Table 5.3.1.a details the distribution of the diacritics in the tested corpus.

Table 5.3.1.a. Distribution of 4,576 diacritics in 4,225 words in the corpus (11,950 words)

| Vowels | without G | in endings |  | G and vowel |  | G without vowel |
|---|---|---|---|---|---|---|
| a | 468 | 284 |  | Ga | 53 |  |

---

[31] The $3^4$ forms don't include the rejected forms with omitted vowels.
[32] The three texts are: http://bit.ly/2fNxD9T, http://bit.ly/2wSk7Wx, http://bit.ly/2vFQbyl.
[33] Texts with such a high rate of vowelization are not rare, particularly in opinion journalism, and even in articles in common newspapers such as in al-Hayat *http://bit.ly/2t10OuQ,* where we found 146 words with diacritic(s) out of 468 words: 156 diacritics are used; 136 words have one diacritic and 10 words have two. *G* is used in 114 words; *–AF*, for indefinite accusative case ending, is used in 31; *–u* is used in 9 occurrences, to mark the active/passive in a verbal form, such as *tuHrj/yustHsn*.

| | | | | | |
|---|---|---|---|---|---|
| u | 414 | 245 | Gu | 29 | |
| i | 120 | 55 | Gi | 43 | |
| F (F, **FA**, FY) | 440 | 440 (95,**339**,6) | GF | 58 | |
| N | 97 | 97 | GN | 5 | |
| K | 210 | 210 | GK | 8 | |
| o | 139 | 84 | Go | 0 | |
| Total | **1888** | 1057 | | 196 | 2492 |

The gemination marker *G* (2,688 occurrences, 59%) is more frequent than all short vowels, nunations and *o* together occurring without *G* (1,888 occurrences, 41%), because it represents a duplication of a bare consonant, thus often referring to another lemma. The most frequent diacritic ending is -*FA* with 339 occurrences, it distinguishes the indefinite accusative from the dual construct state (-*A*, called *mudaf*) form of a noun. The magazine uses exclusively the normative variant of the indefinite accusative -*FA*, as opposed to -*AF*, often used in the *Al-Hayat* or *Annahar* newspapers. Our typographical rules (fathatan alef equiv alef fathatan=YES, Section 4.1) accept both variants. The *o* is a frequent ending because it indicates the dual for nouns or adjectives in order to disambiguate it from plural forms.

Table 5.3.1.b. Lexical coverage of the corpus (11,950 words/5,950 types)

| Missing | Occurrences | Types | | Occurrences (%) | Types % |
|---|---|---|---|---|---|
| Proper nouns | 80 | 38 | | 0.7 | 0.6 |
| Other valid forms | 71 | 26 | | 0.6 | 0.4 |
| Total | **151** | 64 | | **1.3** | 1.1 |

Our algorithm detected in the corpus only one typo error: a bare letter substitution (المغزى/المغرى; ر/ز ; z/r), which indicates an excellent editing quality. The first 15 pages (Syria-Lebanon) were totally covered by our resources except one verb نگّل (*nkGl*,$V62-123) "to torture". The other 167 uncovered occurrences (90 types/5,600) are in the 45 pages from Morocco and may be classified in three categories:

(i) Typo errors, diacritics and glottal stop (16 occurrences): The 4,225 words with one, two or even three diacritics were all validated by our algorithm except 16 words not found in the resources. 11 of them are misplaced occurrences of *G*. Three are true typo errors: the *G* occurs on the wrong bare letter (*tqGSy* instead of *tqSGy*). The other 8 flagged words are cases of inversion vowel-*G* / *G*-vowel. Our typo rules state that *G* must be followed by the vowel. In fact, the two sequences *Ga* and *aG* appear as two glyphs superposed in the same order; they are visually identical, and cannot be distinguished by the editors of *Al-adab*. The rule is observed in 196 cases and there are 8 inversions (*aG/Ga* or *FGA/GFA*).

The 5 remaining flagged "errors" are related to different standards for glottal stop scripting in Morocco and the Levant:
     (a)   بدؤوا / بدأوا ; *bdOwA/bdWwA* (2 occ.) <bdO:aP3mp>
                        ; Morocco/Levant glottal stop rules
     (b)   بمبدأ / بمبدإ ; *bmbdO/bmbdI* (3 occ.)
                         ; <PREP><mbdO:NmsaG>

(a) In Morocco, the suffix –wA (Perfect 3ʳᵈ person masc-plural) at the end of a form is considered as external to the core verb; therefore, the glottal stop rule for the end of a word applies; whereas in the Levant, the suffix is considered part of the core verb; therefore, the glottal stop rule for the middle of a word applies.

(b) Our agglutination grammar rules select the genitive case ending (-i, -K) and in both cases (construct state or indefinite) the glottal stop diacritic followed by i/K should be written *preferably* as I (below *alif*), not as O (above *alif*).

(ii) Proper names (80):
Many proper names were recognized. However, the test collection shows that 80 occurrences (38 types) of proper names were not recognized, representing first names, surnames or place names, that are not included in our lexicon.

(iii) Other forms missing in the lexical resources (71):
The test collection shows that 71 other occurrences were missing in our lexicon, representing 26 types:

- The word *Amazigh* agglutinated or not occurs 27 times.
- The two orthographic variants *tfnAq* or *tfynAq*, denoting the Amazigh alphabet, occurs 12 times.
- The word الهويّاتيّ "identitarian" occurs 16 times as a noun or adjective in the masculine or feminine, agglutinated or not. This word is a derivative with the ending suffixes -yG or -yGap. 11 other occurrences of derived adjectives ending with -yG or _-yGap: إسلامويّ، التلازميّة، الحضريّة، الفلّاحيّة، الرغبويّ، القاعديّة، الموحّديّة، اللسنيّة، راهنيّته، قدحيّة، لاعقلاني;
- 4 nouns (المستفتين، الملالي، شيعًا، الوندال);
- 1 verb (ودستَرها and_*dstara_hA*, "and_put-in-the-constitution_it") .

Morphosyntactic tagging is generally part of a pipeline of written text processing. Unknown words may jeopardize a subsequent deep syntactic parsing of a sentence. Thus, fallback procedures (not implemented) are required to assign a POS to unknown words, such as relational adjectives ending with –yG and typical Arabic proper nouns starting with *Ebd-* or ending with *-Allh* or *-Aldyn*, which are common prefixes and suffixes in Arabic proper nouns.

Summing up, our resource (see our Arabic spell checker http://babelarab.univ-mlv.fr/) has flagged 11 words with partial vowelization: 3 with true errors, and 8 with discrepancies regarding Morocco/Levant standards for glottal stop rules. The fault rate of coverage (Table 5.3.1) in Arabic-Unitex is 1.3%, proper nouns included (0.5%, if excluded), and the fault rate is 1% (0.4 % if proper nouns excluded). Finally, our lexical resources have a better coverage of Levantine usage.

### 5.3.2  An extrinsic evaluation through a local grammar

In the preceding experiment, the system uses information provided in the dictionary: inflected form, POS and inflectional features, and the results are therefore an indirect evaluation of these fields. However, it does not use the lemma field also provided in the dictionary. In this section, we report an extrinsic evaluation experiment devised to assess the system's ability to recognize lemmas.

We made an experiment similar to Traboulsi (2009) and Ben Mesmia et al. (2015) but with our resources. Traboulsi (2009) underlines that *"Despite the fact that the probabilistic approach (the supervised machine learning) and the symbolic approach (the rule based) have been successful in recognizing Arabic person names in news texts, these approaches require large tagged corpora, dictionaries or gazetteers, lists of proper names, which could have been avoided if the local grammar approach was used the way they do." (Section 2)*. Traboulsi recognizes the structure <Reporting_verb><Noun+Human> which is frequent in newspapers. He takes advantage of the frequency of verbs such as *said*, *declared*, *indicated*, … and the predictable occurrence of a subsequent proper noun. To implement his local grammar, Traboulsi uses a cascade of FSTs that apply in a strict order. Ben Mesmia et al. (2015) presented many local grammars for recognizing Arabic Named Entities (ANE) based on a transducer cascade as well. They established word lists, a set of extraction rules based on trigger words and a set of transducers allowing the recognition of several ANE categories.

The advantage of these two implementations is that they dispense with annotated corpora; the drawbacks are: agglutinations are not handled properly, as each possible agglutinated form should appear explicitly in the local grammar, making it unnecessarily overloaded; the word lists are constructed on the fly from the corpora.

Consequently, we expected that, with a rich morpho-syntactic representation, the local grammar approach of these two methods could be adapted to have a better recall/precision. Moreover, it is easier to conceive a local grammar based on a pre-processed, segmented and annotated text. Our rich annotation with lemma, POS and inflexional attribute values helps to craft a more concise and readable grammar. For instance, checking agreement and disagreement between words helps to identify syntactic structures and boundaries, and consequently, semantic slots. Such checks result in more precision in capturing Named Entities.

We built a local grammar (Fig 5.3.2.a) that identifies the verb "to say" in the perfect or imperfect 3$^{rd}$ person masculine singular, followed by a chunk with the noun "minister". The local grammar outputs braces delimiting this pattern, as in:

وقال {وزير المال الفرنسي جان ارتوي} في اثناء الجلسة ان الدوام الجديد

"and_said {minister of_finance French Jean Artuis } (in) during the session

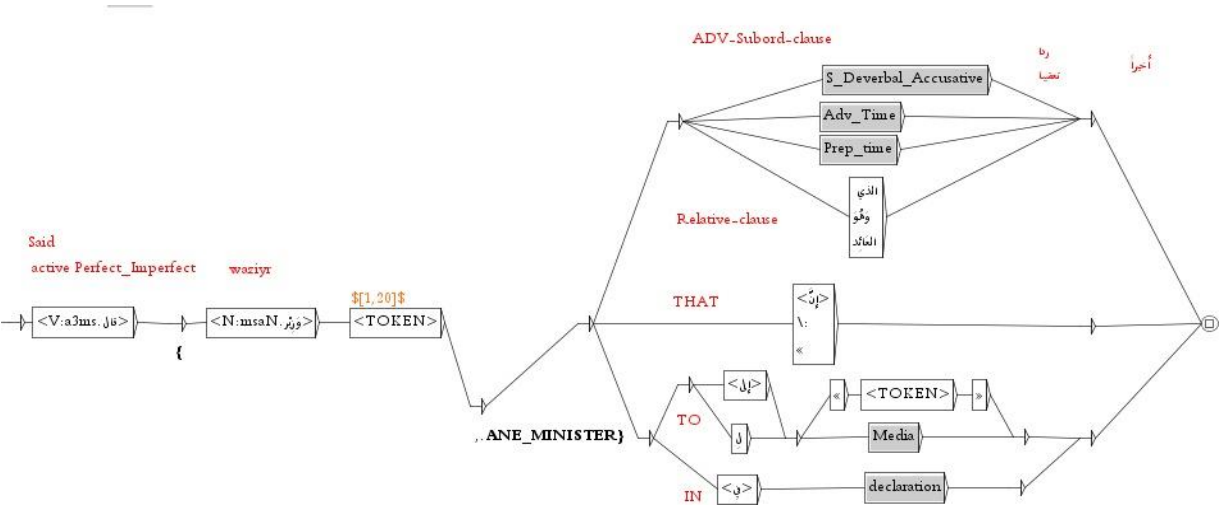

Figure 5.3.2.a. Local grammar identifying ministers

In Fig 5.3.2.a, the box <TOKEN>/$1,20$ defines a window of 20 words in which a pattern indicating the end of the chunk is searched. The local grammar contains 6 graphs and 55 boxes in total. The pattern belongs to one of three types:

- <THAT>: *IinGa* "that" introduces an embedded sentence beginning with a noun. The sentence may also be introduced by a colon or a double quotation mark.
- TO <MEDIA> or IN <DECLARATION>: *IilaY* or *li* "to" may introduce a media slot: journal(ists), Al-Hayat, (press) agency, radio. The preposition *fiy* "in" may be followed by a declaration slot such as conversation(s), conference(s), meeting(s), book(s) as: <مُؤْتَمَر.N:mG>+<جَلْسَة.N:fG>+<حَدِيث.N:mG>+<مُقَابَلَة.N:fG>+<تَصْرِيح.N:mG>+<بَيَان.N:mG>+<لِقَاء.N:mG>+<نَدْوَة.N:fG>+<كَلِمَة.N:fG>+<إتِّصَال.N:mG>+<إجْتِمَاع.N:mG>+<عَرْض.N:mG>+<رِسَالَة.N:fG>+<كِتَاب.N:mG>+<مُدَاخَلَة.N:fG>+<تَعْلِيق.N:mG>+<خِطَاب.N:mG> in the genitive case and either definite, construct state or indefinite, prefixed (or not) by *Al* and agglutinated (or not) to a pronoun such as in "intervention_his" (line 17 in the concordance below)
- ADVERBIAL or SUBORDINATE CLAUSE: It can be "yesterday", "Tuesday" or any date. It can be a relative clause introduced by a relative pronoun or an active participle such as "travelling" or a deverbal noun such as "commenting".

Table 5.3.2. Part of a concordance with 971 matches identified by the local grammar

| | | |
|---|---|---|
| 1 | قال {وزير العدل السوداني السيد عبدالباسط سبدرات,.ANE_MINISTER} | ان |
| 2 | قال {وزير اقتصاد نيكاراغوا نويل ساكاسا,.ANE_MINISTER} | انه |
| 3 | قال {وزير خارجية كوبا روبرتو روبينا,.ANE_MINISTER} | ان |
| 4 | وقال {وزير الاعلام,.ANE_MINISTER} | عقب ال |
| 5 | وقال {وزير الشؤون الاجتماعية أيوب حميد,.ANE_MINISTER} | بعد |
| 6 | وقال {وزير الاعمار الوطني اتيان مبايا,.ANE_MINISTER} | في بيان |
| 7 | وقال {وزير الزراعة,.ANE_MINISTER} | **ردا على اسئلة** » |
| 8 | وقال {وزير الدفاع الاسرائيلي اسحق موردخاي,.ANE_MINISTER} | قبل |
| 9 | وقال {وزير الخارجية الكندي لويد اكسويرثي,.ANE_MINISTER} | في كلمة |
| 10 | وقال {وزير الخارجية البولندي داريوس روزاتي,.ANE_MINISTER} | أمس الثلثاء |
| 11 | وقال {وزير العدل السوداني عبدالباسط سبدرات,.ANE_MINISTER} | خلال |
| 12 | وقال {وزير الخارجية الايراني كمال خرازي,.ANE_MINISTER} | في مؤتمر |
| 13 | وقال {وزير الخارجية البريطاني روبن كوك,.ANE_MINISTER} | بعد |
| 14 | وقال {وزير الخارجية البريطاني روبن كوك,.ANE_MINISTER} | أمس انه |
| 15 | وقال {وزير المال ليم تشانغ يول,.ANE_MINISTER} | في مؤتمر |
| 16 | قال {وزير الخارجية القبرصي اليكوس ميخاليدس,.ANE_MINISTER} | أمس |
| 17 | وقال {وزير الشؤون الاوروبية بيار موسكوفيسي,.ANE_MINISTER} | في **مداخلته** |

We evaluated the recall of the graph on part of ArabiCorpus http://arabicorpus.byu.edu/, an online set of untagged Arabic corpora that contains portions of textual documents from different sources. We have used *Al-Hayat* 1997 (Saudi Arabia).

We launched the search query *qAl wzyr* ("said minister") as a string and we obtained a concordance of 985 occurrences (Table 5.3.2). We discarded the 10 occurrences where *qAl* is a substring of another verb such as *IEtqAl* "arrested" or *IstqAl* "resigned". The remaining 975 are the target of our local grammar.

The local grammar identifies 971 occurrences (see Table 5.3.2) of the entity {MINISTER} out of 975 (99,6% recall). The 4 missing occurrences contain:

- One occurrence of *O* (instead of *I* or *A*) in *IinGa*, which is a spelling mistake since reporting verbs should be followed exclusively by *IinGa*. Our grammar identifies vowelization variants of the lemma <IinGa> (such as *In, Iin, InG, An, AnG, AnGa*, etc) but not of the lemma <OanGa>.
- One occurrence of *radGAF* "responding", tagged as unknown word. The lemma of this deverbal noun is missing in our dictionary (see concordance, line 7): *radGAF* is a deverbal noun based on a simple verb ($V31 to $V36 in our encoding, Neme 2011); these deverbal nouns are irregular.
- One occurrence of the pattern *Ily Al-SHAfyGyn* "journalists". This noun has two pronunciation variants *SuHaAfiyG* and *SaHaAfiyG* (cf. Section 4). In our lexicon, we opted for *SuHaAfiyG* and did not encode the variation, whereas in our grammar (cf. Fig 5.3.2.b), we used <**Sa**HaAfiyG> as lemma to identify the inflected forms.
- One occurrence without any of the patterns recognized by the grammar to locate the end of the chunk. The contents of the declaration are before the verb "say" and the sentence does not mention the media: "Will they find it…, as said the previous American minister of foreign affairs Warren Christopher?"

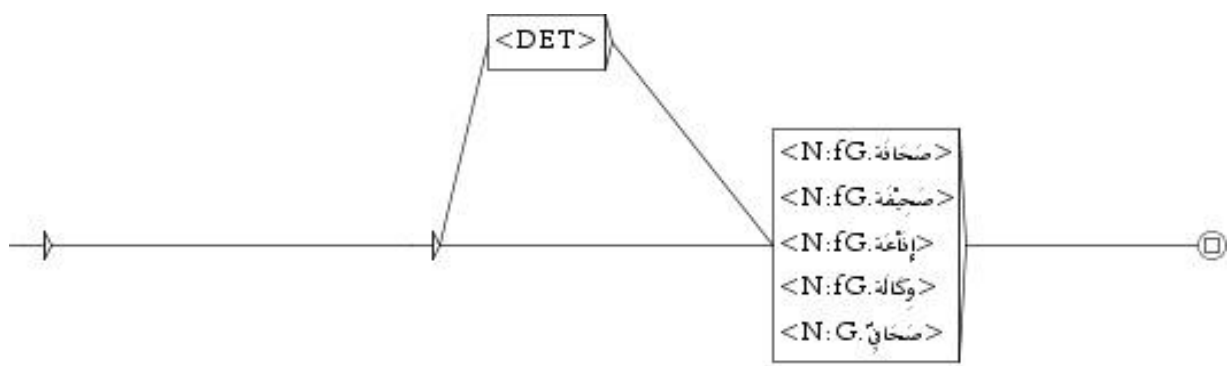

Fig. 5.3.2.b. The subgraph <Media> in the local grammar of Fig. 5.3.2.a

The use of very informative lexical resources also facilitates the manual construction of local grammars. In the lexical resources, the lemma <journalist> has 54 inflectional variations. In a local grammar, <journalist.N:G> recognizes 18 forms in the genitive case and excludes the 36 other variations. This representation identifies standalone forms, but also agglutinated forms with *Al* or with 12 potential pronouns. Furthermore, it is useless to represent in the local grammar (Fig. 5.3.2.b) the agglutinated pronoun <PRO>, since the result of morphological analysis represents any variation of <journalist.N:G> separately, even before a possible agglutinated pronoun (see line 17 in the concordance). For computational linguists, such simple and natural formalization of the local grammar represents an enormous gain and simplification.

Likewise, all the inflection of a verb may be covered by a lemma followed by inflectional features such as <say:a3ms>, with unspecified tense, and thus referring to both active perfect and imperfect. Moreover, since the segmentation of words is handled by our agglutination grammars, agglutinated forms with proclitics such as "and said" and optional enclitic pronouns may be detected simply by the formal representation <say:a3ms> (<E> + <PRO:3s>) which retrieves "said" and "said it". This turns local grammars more readable.

As we have said above, an adverbial clause may constitute the pattern that indicates the end of the chunk. An adverbial clause may begin with a deverbal noun such as "commenting", generated automatically (with 10 000 other deverbal nouns) from an augmented verb (classes $V61-$V70). From `EqGb,$V62-123`, "to comment", we have generated a dictionary entry for the lemma `taEoqiyb`, from which the indefinite accusative form تعقيباً is generated as an inflected form and encoded as a line in our lexicon as `taEoqiybaAF,taEoqiyb.N+`**`Masdar=EqGb`**`:msiA`. But the native linguist may extend this vocabulary in the local grammar by adding synonyms of "commenting" such as منتقداً "criticizing", based on *introspection*, even if the synonyms do not appear in the corpus.

To conclude, the ability to recognize lemmas and their variations is tested successfully. Our resources allow for helpful conciseness in the detection of inflected forms by local grammars. Moreover, they make it unnecessary to tag corpora since we tag texts automatically using a dictionary which covers more than 76 000 lexical entries[34]. Besides, they allow to annotate corpora semi-automatically as an input for supervised learning.

## 5.4 Arabic-Unitex versus BAMA lexicon

Many features distinguish Arabic-Unitex from the BAMA lexicons. Here is a survey of the main differences and similarities.

a) Usage in the Levant

Arabic-Unitex is mainly based on the Levantine usage of Arabic language. The Levant defines *de facto* the Modern Standard Arabic usage. This tradition dates back to when the Umayyad caliph Abd Al-Malik made Arabic the official language during his reign (685-705) in Damascus. In Arabic-Unitex, most lexical entries are citation forms attested in paper modern dictionaries printed in Beirut after 1970: Abd-Nour (2006), Khalil Al-Jar (1973, Larousse) and others; we used https://www.almaany.com/ to double-check meaning and usage. We also included terms and neologisms found in the Arabic Wikipedia, the Nemlar corpus, and the *Annahar* (Beirut) and *AlHayat* (KSA-Beirut-London) newspapers.

---

[34] The list of proper nouns (around 6000) includes name of countries and important cities, Arabic and foreign forenames and family names such as celebrities: Ronaldo, Rif(v)aldo, B(P)edro Almodovar, and George Bush, etc. This list was created first by extracting the proper nouns from the Nemlar corpus. Secondly, we processed many newspaper corpora, short novels and other modern fictions with our Arabic-Unitex resources. From the unknown words list output by the Unitex tagger, we extracted the simple and agglutinated forms of proper nouns. The proper nouns represent often more than half the unknown-word list. We encode them manually and such encoding enables recognizing agglutinated proper noun forms such as <CONJC><PREP><NPr>.

The BAMA lexicon is derived from the ALPNET project, and based mainly on Hans Wehr's bilingual dictionary[35] (1952). BAMA includes Egyptian variants such as *kabuwriyA* كبوريا "crab", *mavGaAl* "sculptor" مَثَّال; *miMolawoz* مشلوز "apricot", excluded from Unitex, which contains instead *saloTaEuwon (and saloTaEAon)* سلطعون; *naHGaAt*, نحّات; *miMomiM* مِشْمِش, which are all in BAMA, as well. BAMA also includes old terms such as *jazuwr, niyb, ZaEuwn,* (resp. "fat camel for butcher meat", "old female camel", "load camel").

b) Loan words

Both BAMA and Unitex include the standard Babylonian naming of months such as *Oayoluwol* أيلول, current in the Levant and the Gulf, and both lexicons also include the names borrowed from English such as *September,* current in Egypt, Sudan and Libya. Neither lexicon includes the denominations of French origin such as *Janvier*, in use in Tunisia, Algeria and Morocco[36]. The month names used in the Islamic lunar calendar for religious events and ceremonies are included in both lexicons.

Both BAMA and Unitex include loan words: *dakotuwr, bruwfisuwr, bruwtiyn* "doctor, professor, protein". BAMA lists both variants *bridoj* and *briydoj* "bridge (game)", while Unitex inventories only the second representation. We *preferably* represent the vowel with the bare letter *y*, in keeping with the current tendency to write loan word vowels with bare letters.

c) Verbal inflection

In BAMA, we counted 415 perfect passive stems, 2845 imperfect ones, 116 stems for the imperative mode, and no energetic mode. Active and passive participles are described in the BAMA lexicon not as inflections of verbs but as adjectives and nouns. In Unitex, we have covered them as inflected forms for 15400 verbs. Note that the passive mode is possible for intransitive verbs such as *niyma bi_Al-firaAMi* "(it) was-slept in-the-bed". Contrariwise, Unitex covers some adjectives in the form of participles, e.g. *MaAeiE* شائع "current". This flaw needs to be fixed, at least for common adjectives.

d) Lemmas with suffixed plurals

In paper dictionaries, some lexical entries are in the plural, because the correspondent singular form exists with another meaning. In our inflectional approach, the lexicographer may encode a citation form in the plural. In Unitex, some lexical entries are lemmatized in the plural, e.g. `qalawiyGAt,N0aAt-p-0`, "alkali (chemistry)". The singular is an adjective. The noun *DaruwriyGaAt* ضروريّات means "necessities"; its singular counterpart is used only as an adjective meaning "necessary". They are encoded as independent lemmas in Unitex:

```
DaruwriyGaAt,N0aAt-p-0/       ضَرُوريَّات
DaruwriyG,A0000-g-uwna/       ضَرُوريّ
```

In BAMA, both lexical entries are encoded with the same lemma `Daruwriy~_1` but with different POS:

```
Daruwriy~_1    Drwry    Daruwriy~    N-ap    necessary/requisite    Daruwriy~/ADJ
Daruwriy~_1    Drwry    Daruwriy~    NAt     necessities            Daruwriy~/NOUN
```

---

[35] The original edition is in Arabic-German "Arabisches Wörterbuch (1952)", published later in bilingual Arabic-English edition as "A Dictionary of Modern Written Arabic".
[36] Unitex should include all these month denominations with features indicating the region of usage +Levant, +EgSuLy, +Maghreb, respectively for the names of Babylonian, English, French origin.

The singular of *mudaAEafaAt* "consequences" is *mudaAEafap* "the doubling", encoded in Unitex by:

```
muDaAEafap,N00ap-f-At/    doubling       masdar+DAEf   مُضَاعَفَة
muDaAEafaAt,N0aAt-p-0/    Consequences                 مُضَاعَفَات
```

In BAMA, both lexical entries are encoded with the same lemma `muDAEafap_11` and the same POS:

```
muDAEafap_1    mDAEf    muDAEafNapAt    doubling/compounding    muDAEaf/NOUN
muDAEafap_1    mDAEf    muDAEafNAt      complications           muDAEaf/NOUN
```

BAMA contains two variants of "sixties", encoded in two lemmas whereas Unitex contains only the first variant:

```
    (a) sitGiyonaAt,N0aAt-p-0/          سِتِّينَات    sixties (in BAMA)
    (b) sitGiyoniyaAt,Not-in-Unitex/   سِتِّينَات    sixties variant (in BAMA)
```

We checked the usage of both variants through the Arab countries (Egypt, Syria, Kuwait, Jordan, Morocco) in a corpus of newspapers taken from arabiccorpus.byu.edu. The corpus has 3046 occurrences of (a) and 2093 of (b). We did not identify any difference in meaning or usage between the variants. Both are used almost at the same frequency in these newspapers, except for *AlHayat* 1997 (1031 a, 18 b) and 1996 (1198 a 23 b). It seems that *AlHayat* has a strict editorial policy and uses almost exclusively the (a) variant. Since there is no difference, we decided to create a new inflectional transducer that generates the –yaAt variant beside -aAt but attaches both to the same lemma (a) `sitGiyonaAt,N0_y_aAt-p-0`. We re-encoded similarly all this family of words: "twenties, thirties, …".

We have almost 200 lexical entries with –aAt suffixed plurals; this list need to be completed.

e) <u>Broken plurals</u>

BAMA includes two lemmas for *xaTar/OaxTaAor/maxaATir*:

```
xaTar_1      >xTAr      >axoTAr      N       dangers    >axoTAr/NOUN
maxATir_1    mxATr      maxATir      Ndip    dangers    maxATir/NOUN
```

whereas Unitex considers both BP forms as inflections of the same lemma (Neme & Laporte, examples 149-151):

```
    xaTar,$N300-m-FvEvL-OaFoEaaL-123/     أخطار   خَطَر
    xaTar,$N300-m-FvEvL-FaEaaLiB-m123/    مخاطر   خَطَر
```

## 5.5 Drawbacks and possible improvements

Since the breakthrough of the BAMA lexicon (Buckwalter, 2002), the majority of new scientific papers on Arabic NLP relies on this lexicon and on its related algorithm, *"a de facto standard tool which is widely used in the Arabic NLP research community"* (Attia et al, 2011).

Attia et al. (2011, Section 2.1) also point out the drawbacks of BAMA; nevertheless, no viable and better alternative has been proposed so far. *"After all aspects of morphological analysis have been adequately addressed, the only way to improve the quality of the analysis is by improving the lexicon."*(Buckwalter, 2007, 3.6 Lexicon Design and Maintenance). ***Improving the lexicon for Buckwalter may be done by enhancing the lexical coverage and by increasing the level of grammatical detail***. He advocates an enhancement of BAMA (2004) by inserting traditional labels (Buckwalter, 2007, section 8):

- **gender, number**, humanness (for noun)
- **active and passive participles** and verbal nouns, deverbal noun (masdar from simple form or **augmented form**) **(**cf. Section 3.4**)**
- elative such as "bigger/the-biggest"
- instance noun, unit/collective noun
- verb features such as transitive, intransitive, grammatical colocations.

We do agree with the mentioned improvements. Our proposal of a new approach to Arabic morphology involves the *pattern-and-root model*, and a large and contemporary lexicon. Our alternative to BAMA is entirely based on the Semitic tradition, one fully inflected lexicon (lemma-based), the pattern-and-root model, and a look-up procedure in the fully inflected lexicon. Most of the enhancements recommended by Buckwalter (in **bold**, cf. 5.2) are included in Arabic-Unitex from its inception. The elative such as "bigger/the-biggest" was encoded for almost 200 adjectives and needs to be extended. Instance nouns, also called cognate nouns, such as ضربة *darb_ap* "hit_one", and unit/collective nouns such as نمل/نملة *namlap/namol*, "aunt_one/aunt_collective" are part of the lexicon and need a systematic encoding in Arabic-Unitex. Arabic-Unitex needs exposure and more testing by applications in order to be further validated.

# 6 Compression

The Unitex programs were adjusted in 2010 to Arabic morphology in order to handle:
- Semitic inflection and infixes,
- proclitic and enclitic agglutination,
- partial vowelization.

In the standard Unitex process, an inflected-form dictionary is compressed into a minimal acyclic deterministic finite automaton data structure in order to be stored in RAM for fast retrieval (Revuz, 1992).

## 6.1 The compression algorithm

The input of the Unitex dictionary to the compression algorithm is a text file whose lines are of the form:

```
<inflected form>,<lemma>.<grammatical:inflectional-codes>
```

like, for example:

```
        takotubu,ktb.V:aI3fsN    / compact tag:  __246.V:aI3fsN
        xawanapN,xaAoein.N:qiN   / compact tag:  __01Aoei4.N:qiN
                                 / خَاْئِنِ BP of خونـةٌ  /
```

The compressed version of the dictionary is a finite state transducer that associates each inflected form with its lemma and codes. The algorithm spares space to store the inflected forms by representing the transducer in the form of a Minimal Acyclic Deterministic Finite Automaton. In order to minimize the space needed to represent the lemmas and codes, it replaces them with a compact tag that contains enough information to restore the complete entry from the inflected form. The standard version of the algorithm, applied to the entry `looks,look.V:P3s`, for example, produces the compact tag `1.V:P3s`. At lookup time, the inflected form `looks` is known, and the lookup program can rebuild `look.V:P3s` from the compact tag `1.V:P3s` by interpreting it as "remove 1 letter from the end of the inflected form and add `.V:P3s`". This strategy is very effective for many languages because it takes advantage of the regularities of the language's inflection system. For English, almost all entries for the third person of the present share the same compressed code "`1.V:P3s`" since the third person of the present of almost all verbs is the infinitive form plus *s* at the end.

However, the nature of Semitic languages makes this suffix-based approach very ineffective. The strategy of our Semitic-oriented version of the algorithm consists instead in indicating which letters from the inflected form should be kept to restore the lemma. Given the inflected form `takotubu`, the `246` substring in the compact tag (above) means that we need to keep the letters `#2 (k)`, `#4 (t)` and `#6 (b)` from the inflected form to obtain `ktb`. In case some letters are missing from the inflected form they are added in the compressed form. For instance, if we have the inflected form `xawanapN` and the lemma `xaAoein`, we compress it as `01Aoei4` which means: letter `#0 (x)`, letter `#1(a)`, followed by the substring `Aoei` and the letter `#4 (n)` from the inflected form to obtain `xaAoein`.

In order to produce compact tags that are more likely to be shared by other entries and thus improve the compression rate, the algorithm tries all possible compact tags and keeps one that maximizes the number of letters copied from the inflected form. For instance, if we have the infected form `abcdefgh` and the lemma `hbc`, we could represent it with several codes: `hbc` (no letter copied from the inflected form), `7bc` (h copied from the end of the inflected form and adding `bc`) and `h12` (adding h and then the 2 letters `bc` copied from the inflected form). Our heuristic will select `h12` because it reuses two letters from the inflected form.

ADJUSTMENTS TO DICTIONARY LOOKUP IMPLEMENTATION

We adapted the Unitex dictionary lookup procedure to this Semitic-oriented compression strategy. Moreover, we adapted the lookup procedure so that it is *tolerant to partial vowelization and other Arabic typographical rules* (cf. Section 6.3). Our version finds for each input word (without vowels, partially or fully vowelized) those candidate forms compatible with the input word. When a diacritic is present in a surface form, the lookup procedure retains the candidates with the same diacritic at the same position in the compressed dictionary.

We also equipped the lookup procedure with *a hash table data structure stored in RAM memory*, which avoids to repeatedly search the minimal acyclic deterministic Finite State Automaton (MADFA) for occurrences of the same word. The procedure looks up the word in the hash table first; if it does not find it, it searches the MADFA and stores the entry in the hash table, in anticipation of other occurrences in the text. This speeds up the lookup by almost 50 times. This feature is independent from the compression strategy and has been adopted as the standard Unitex lookup. [37]

In addition, we pass the agglutination grammars to the lookup procedure in the form of a *flattened* FST. Each agglutination grammar is manually produced in the form of a network of graphs and subgraphs, which are compact, readable and reusable. Flattening replaces calls to subgraphs by copies, taking advantage of the fact that the network is not recursive. The global flattened grammar (grouping verbs, noun/adjectives and particle agglutination grammars) consists of 1 graph with 60 states and 286 transitions, instead of 25 graphs and subgraphs, totalling 175 states and 369 transitions. As a result, the flattened FST makes lookup approximately 2 times faster for the price of a simple compilation[38].

## 6.2 Two compression experiments

The full-form dictionary has 6 million surface forms. It is 340 Megabytes in plain text in Unicode UTF-8.

With the Semitic-adjusted version, we compress it into 13.5 Megabytes. The compilation of the 1,150 inflection graphs and 4,000 subgraphs takes one minute. The generation of the 6 million forms takes 10 seconds; the compression and minimization of the full-form lexicon takes one minute on a Windows laptop[39]. The morphological analysis processes almost 1000 words/second or 3 pages/second for vowelized or unvowelized text alike.

The compression ratio is better (see Table 5.4), and the lookup much quicker, if we compress separately the entries inflected in the Semitic mode. We have split into two parts the dictionary of 76,000 lemmas: 19,600 ones with inflection in the Semitic mode and 56,400 ones with inflection in the concatenative mode or no inflection.

From the 19,600 lemmas with Semitic inflection, we have generated 4,280,000 forms and a 228-Megabytes flat file. The Semitic-oriented version of the compression algorithm produces a 10.5-Megabyte compressed file.

From the 56,400 lemmas with concatenative inflection, we have generated 1,805,000 forms and 114 Megabytes flat file; the standard compression algorithm produces a 0.5-Megabytes file.

---

[37] Wintner's morphological analyser of Hebrew implemented in Java also stores the Hebrew lexicon in a lookup table (Wintner, 2008, Section 2.2): *"contemporary computers can efficiently store and retrieve millions of inflected forms. Of course, this method would break in the face of an infinite lexicon (which can easily be represented with FST), but for most practical purposes, it is safe to assume that natural language lexicons are finite."* Indeed, if the hash table approach were applied to an Arabic lexicon with all partially vowelized forms, the list would grow to an estimated tens (or hundreds) of billions of forms, almost unmanageable for a lookup table.

[38] Unitex includes a "compile and flatten" variant of the compiler for transducers. The output of Unitex transducer compilation is in the FST2 format. The basic version of the compiler "*conserves the architecture in subgraphs of the grammars, which is what makes them different from strict finite state transducers. The Flatten program allows you to turn a FST2 grammar into a [single] finite state transducer whenever this is possible, and to construct an approximation if not. This function thus permits to obtain objects that are easier to manipulate and to which all classical algorithms on automata can be applied.*" (Paumier, 2016, UNITEX-User manual 3.1RC, Section 6.2.2)

[39] Windows 7, HP Zbook 15 G2, i7- 250GHz x64, Memory: 16 GB.

Table 6.2. Comparing the two experiments of compression

|  | Together | Separately | |
| --- | --- | --- | --- |
| Compression algorithm | Semitic | Semitic | Concatenative |
| Number of entries | **6 082 374** | **4 280 000** | **1 805 000** |
| Flat File Size (Megabytes) | 341 | 228 | 114 |
| Bin file size (Megabytes) | **13.5** | **10.5** | **0.5** |
| INF entries | 83 858 | 65 337 | 2 859 |
| States | 252 774 | 200 450 | 30 746 |
| Transitions | 586 103 | 427 027 | 68 305 |

With these two compressed files, the analysis speeds up to 1,800 words/second on a 2014 Windows laptop (5000 words/second on MacBook Pro i7, 2,0 GHz, 8 GB RAM), which is almost three times the speed of AlKhalil-2 (632 word/s) or BAMA (685 words/s). Compared with the compression with the Semitic compression only, the split speeds up the analysis by 80%.

In Neme and Laporte (2013), we compare the performance of our parser and MAGEAD-Express (both analysers cover verbal inflection and use FST technologies):
- *The resources of MAGEAD-Express (8700 verbs) compile in 48 h, and the analysis of a verb takes 6.8 ms* (Altantawy *et al.*, 2011- Octobre:123) (Section 2.4.2)
- *Neme (2011-August) describes a morphological analyser for Arabic verbs with a comprehensive lexical coverage:15 400 verbs. The dictionary compiles in 2 minutes and the analysis of a verb takes 0.5 ms on a 2009 Windows laptop, outperforming MAGEAD-Express* (Section 2.4.5)

With Hebrew resources (21,000 lemmas/0.5 million forms), Wintner (2008) reports the following numbers when using an FST lookup procedure and compression: 25 minutes to compile and compress the resources; and the analysis speed is 83 words/second. On the other hand, with the same Hebrew resources, when using a lookup with a hash table and a Java classical programming platform, the compilation of the resources takes few seconds and the analysis speeds up to 1500 words/second.

Our lookup is fast because the design is simple. Our inflectional ALR has a solid, straightforward Arabic morphological basis which made it possible to generate a comprehensive, detailed, accurate full-form dictionary, including literal morpho-phonological variants and with vowels fully represented. No on-the-fly computation of morphological changes in agglutinated forms is required during the analysis. The agglutination grammars in the ALR specify literal orthographical variants, which also speeds up the process.

## 6.3 Algorithm for restoring vowels

As explained before, the compressed dictionary consists of a transducer containing all possible fully vowelized forms. The lookup procedure explores in parallel the transducer and the text to find matches. Once a match is found, the transducer gives access to a compact tag that can be used to reconstruct a full dictionary entry.

The transducer/text matching takes into account partial vowelization and other Arabic typographical rules. The rules enabled by the user in the configuration file (see Section 4) affect this matching process. The code that explores the transducer looks first for an exact match but also looks for alternate matches depending on the rules that have been activated[40].

For instance, with the predefined rules, if the dictionary contains the form *kitaAbFA*, the lookup procedure matches *ktAbFA* in the text and restores the missing vowels from the dictionary. It also matches the input forms *ktAbAF* and *kitAbAF,* if the rule about the inversion between *A* and *F* is active. Then it uses the compact tag associated to *kitaAbFA* to get the lemma *kitaAb* and the POS/inflectional codes `N:msiA`. In the end, the output (cf. Fig.5.3) contains the following line with the fully diacritized form retrieved from the dictionary:

```
kitaAbAF,kitaAb.N:msiA
```

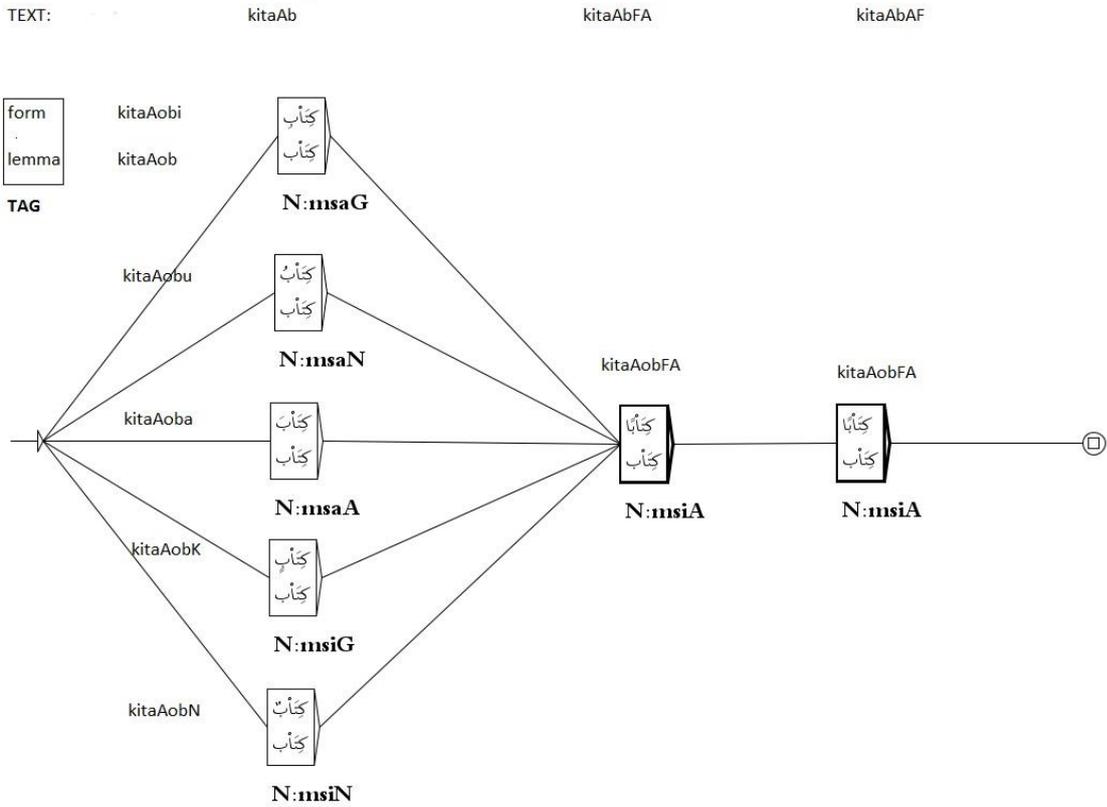

*Fig. 6.3. Restoring the vowels. Parsing outputs of the sequences* kitAb, kitaAbFA, kitaAbAF

---

[40] As a preprocessing, we normalize the text by keeping one space between words and trimming the *tatweel* character from words. This character is used for text justification and to extend the horizontal connexion line between two connected letters, as in *kt__Ab* كتــاب instead of كتاب. Obviously, the *tatweel* is not used in the dictionary.

## 6.4 Exploiting lattice output in an NLP pipeline

As opposed to most taggers, who output a single analysis for each word, our tagging outputs several analyses, forming a lattice. In this section, we show how such a labeled word lattice can be exploited, either with Unitex or by including it into a processing chain with other systems.

First, Unitex itself can search a labeled lattice for a user-defined query, as exemplified in Section 5.3.2.. The presence of several analyses in parallel in the lattice might theoretically reduce the precision of the search results. However, this kind of lattice search is probably the most popular case use of Unitex in academia and NLP companies, since the Unitex default pre-processing looks up a compressed dictionary and provides the list of possible tags for each word; and, with typical queries, precision is not significantly lower than with a search performed on classical, single-analysis tagged text. Fairon, Paumier et Watrin (2005) quantify the difference in precision on the recognition of French syntactic structures. They formalize the syntactic structure of French verbs in order to generate "*parametrized graphs (Unitex, User Manual 3.2 Chap. 9), drawn with the help of Recursive Transition Network (RTN) formalism. Such graphs describe linguistic constructions [...]. [The] method does not distinguish between pattern matching and parsing. Once we have generated graphs, we consider them as patterns. We use the pattern matching function of Unitex to find all matching sequences in a text. If sequences are matched by a graph, then we can say that we have parsed these sequences*". They make an evaluation of the identification of the syntactic structures for the most common five verbs in a corpus of 1.5 Million tokens. They demonstrate that the ambiguities present in the tagged lattice output do not prevent the syntactic parsing of verbal constructions and reach a comparable precision whether applied to an input with lattice ambiguity or without by using a statistical approach, like the one in TreeTagger.

Second, the labeled lattice can be turned to tagged text by selecting a path. Krstev et al. (2018) do that with Unitex for Serbian text without diacritics. We summarize their pipeline by the following:

1. For each word $W_b$ they retrieve all possible Serbian words that use diacritics.
2. For each word $W_b$ they rank all the possible candidates ($W_{b1}, W_{b2}, \ldots, W_{bn}$) according to the possibility of their occurrence in a text.
3. For each word $W_b$ that has more than one possible candidate $W_{bi}$, their procedure uses heuristics (based on the Corpus of Contemporary Serbian and processed for uni-, bi- and tri-gram frequencies), lexicons and rules (local grammars) to choose one.

"*The evaluation results reveal that, depending on the text, accuracy ranges from 95.03% to 99.36%, while the precision (average 98.93%) is always higher than the recall (average 94.94%)*" (Krstev et al., 2018:41)

Similar experiments have already been tried with success with a discriminant model or a hidden Markov model on lattices obtained with dictionaries and other tools than Unitex, in Turkish (Sak et al., 2011) and in Arabic (Chennoufi, Mazroui, 2017 ).

Sak et al. (2011) select the most likely analysis via a discriminative algorithm by exploiting the morphological tags associated to agglutinated morphemes in a Turkish token, "*The problem of finding the most likely morphological analyses of the words in a sentence can be solved by* *estimating some statistics over the parts of the morphological analyses on a training set and then choosing the most likely parse output using the estimated parameters. For parameter estimation, we use the averaged perceptron algorithm.*"

They conclude that *"Morphology is a very important knowledge source for morphologically complex languages like Turkish. Using these resources and tools, one can parse a text corpus and obtain the morphological analyses of the words as well as their probabilities, disambiguate the parse outputs, train statistical models using the web corpus, and build applications that fully exploit the information hidden in the morphological structure of words."*

Chennoufi & Mazraoui (2016) present a solution with HMM modeling for a diacritizer that uses *"a hybrid system for automatic diacritization of Arabic sentences combining linguistic rules and statistical treatments"*. The processing is divided into 4 stages, and the 4$^{th}$ stage is a fallback procedure for unknown words:

*"After morphological analysis step that gives for each word all its possible diacritizations, and following the validation step of transitions between pairs of diacritized words and the application of diacritic rules, we present the third stage of diacritization process. It consists of a statistical treatment based on the hidden Markov models and the Viterbi algorithm (Neuhoff, 1975), which provides the most likely diacritized sentence (Fig. 2). The representation of observed states of HMM are the Arabic words without diacritics (eg " فهمتم " /fhmtm/) and the hidden states are diacritized word forms (eg " فَهِمْتُم " /fahimotumo/) (Elshafei et al., 2006; Bebah et al., 2014). This model states provided the best scores of automatic diacritization compared to other hidden states like lists of diacritical marks (Bebah et al., 2014).*

They conclude, *"The good performances of our system are consequences of:*
- *The robustness of the **second version** (with a large improvement of lexical coverage compared to the first one) of AlKhalil analyzer used by our system in the morphological stage;*
- *The use of syntactic and diacritic rules;*
- *The strong representation of the corpus used in the training phase given its large size."*

Summing up, even if they output a labeled lattice with several analyses in parallel, our linguistic resources will improve downstream Arabic NLP pipelines, because the lexicon has comprehensive coverage and *unknown word*s may easily be added to the lexicon with their inflexional variations; moreover, specific symbolic grammar rules or statistical approaches may be also applied to remove paths from the lattice outputs, and with its fine-grained grammatical tags, our approach can enhance further the accuracy of statistical algorithm processing in the future.

Our resources-centered approach to Arabic NLP with Unitex reinforces the readability and maintainability of lexica and grammars for Arabic speakers and linguists; combined with machine learning, it can improve upon the best hybrid solutions in the current state-of-the-art in Arabic NLP.

# 7  Conclusions

Why do computer scientists ignore vowels in their Arabic-processing systems? As Maamouri et al. (2006) note, *"Since non-diacritized text prevails, the Arabic NLP community seems to have accepted using it as the de facto 'real world' information material without feeling an obligation to question its choice/use, even espousing the idea sometimes that the robustness of software algorithms can deal with the problem and reduce the negative effect of the missing information on their research."* […] *"The prohibitive cost and the usually unequal and questionable quality of human/manual diacritization have led the scientific Arabic NLP community and its sponsors to focus more on volume of un-vowelized data so far."*

Also note their excellent later discussions presented in *Diacritization: A challenge to Arabic treebank annotation and parsing* (Maamouri et al. 2008): *"Much parsing work with the ATB has used the unvocalized form, on the basis that it more closely represents the "real-world" situation. We point out some problems with this usage of the unvocalized data and explain why the unvocalized form does not in fact represent 'real-world' data"*. The fact that vowels are largely absent from written text does not prevent us from taking advantage of them in applications.

Contrariwise, our system presents two dozen rules handling short vowels and gemination omission and glottal stop variations, each of which may be enabled or disabled according to the goal of the application. As in traditional dictionaries, we also provide lexicographers with a simple means to represent short vowel variations in inflected forms, grouping more forms under the same lemma. We have implemented as well adequate and specific inflectional operators that can be used easily by native linguists in Arabic (and Austronesian languages).

Our approach to Arabic morphology redefines and reuses standard concepts from the Semitic tradition (Neme & Laporte, 2013). Our lemmatized representation and implementation of morphology is similar to the grammatical tradition in that prefixes and suffixes of verbs are included in the inflectional representation and we account for clitics independently in agglutination grammars; whereas in the implementation of the stem-based approach, the boundaries between such affixes and clitics are ambiguous and fuzzy. Our distinctive approach to morphological analysis is integrated in a ***one-step processing***. This processing is defined by the application of agglutination grammars that validate the delimited word forms (DWF), which includes checking a core POS represented by a diacritized full form, and selecting only compatible solutions when the DWF is partially vowelized.

The supervised machine learning approach requires a large tagged dataset in order to be successful (for instance in Named Entity Recognition). Such resources are scarce for Arabic, or at least difficult (repetitive and "tedious") to tailor to specific needs. Contrariwise, with our lexical resources (once validated thoroughly) and a local grammar approach, such dataset resources are unnecessary or can be produced semi-automatically.

The excitement (2000-2018) for exclusive Machine Learning and statistical approaches comes mainly from the fact that the market needs quick development of viable solutions. Such solutions in simple applications, such as spell checking, indexation…, have satisfactory accuracy for English and even French, but not for Arabic. Previous experiences with ML (till 2017) show that these approaches were not able to propose satisfactory and accurate solutions, even in simple applications. Statistical approaches reached their limits for Arabic NLP, as is demonstrated by the superiority of the Microsoft Arabic spell checker, based on lexical resources, over the one in GoogleDocs. Without Arabic lexical resources, the output of an NLP pipeline is disappointing.

Even with the latest RNN-LSTM technologies, recent publications show that using a rich morphological analyser with large coverage will improve *drastically* the accuracy of morphological tagging. In the case of Arabic NLP, it is time to take the best from all fields of NLP and linguistics: lexicography, morpho-syntactic rules, FST technologies, semantic methodologies, and statistical approaches.

The Arabic-Unitex resources provide a lexical coverage of 99 percent of the words used in online news media, and they offer an integrated, simple and efficient way of restoring vowels in partially vowelized or unvowelized words, by using almost standard finite-state technologies and algorithms. Moreover, we have tested our encoding scheme with native linguists, without noticing any strain in the learning process. Arabic-Unitex complies at the same time with the Semitic tradition, lexicographic tradition, a straightforward legibility and incrementability of the resources.

# 9 Appendixes

The Unitex predefined Arabic typographical rules are the following:

```
fatha omission=YES                   /a
damma omission=YES                   /u
kasra omission=YES                   /i
sukun omission=YES                   /o silent vowel
superscript alef omission=YES        /R superscript  alif
fathatan omission at end=YES         /F
dammatan omission at end=YES         /N
kasratan omission at end=YES         /K
shadda fatha omission at end=YES     /Ga
shadda damma omission at end=YES     /Gu
shadda kasra omission at end=YES     /Gi
shadda fathatan omission at end=YES     /GF
shadda dammatan omission at end=YES     /GN
shadda kasratan omission at end=YES     /GK
shadda fatha omission=YES
shadda damma omission=YES
shadda kasra omission=YES
shadda superscript alef omission=YES  /R in AllGRhu = Allaah
solar assimilation=YES               /insertion a gemination after consonant
lunar assimilation=NO                /no  assimilation  exclude  assimilation
                                     /after non-coronal consonnant
Al with wasla=YES                            /L   Al =>Ll
alef hamza above O=YES                       / O => A
alef hamza below I to A=YES                  / I => A
alef hamza below I to L=YES                  / I => L
fathatan alef equiv alef fathatan=YES             /at the end FA => AF
fathatan alef maqsura equiv alef maqsura fathatan=YES   /FY =>YF
```

*Table 5.1.b. Inflectional features and values carried by POS used in Arabic-Unitex*

| POS carrying the value | FEAT:VALUE | In English | Encoded example | In Arabic | Arabic examples |
|---|---|---|---|---|---|
| <V>,<N>,<A>,<PRO> | Gender | | | | |
| | :m | masculine | <PREP><N:fsDA><PRO:Gen:3fs> | مذكّر | لرجلها |
| | :f | feminine | <DET><N:fsDA> | مؤنّث | الشمسَ |
| <V>,<N>,<A>,<PRO> | Number | | | | |
| | :s | singular | <N:msiN>, <N:msiG> | مفرد | قائدٌ، خائنٍ |
| | :d | dual | <N:fdiN>, <N:mdiA> or <N:mdiG> | مثنى | طاولتان، مراقبَين |
| | :p | suffixed plural | <N:fpiN>, <N:mpiN> | جمع سالم | طاولاتٌ، مراقبون |
| <N>,<A> | :q | broken plural (non-suffixal) | <N:qiN>, <A:qiG> | جمع تكسير | قادةُ خَوَنةٍ |
| <N>,<A>,<V:F>,<V:M> | Definiteness | | | | |
| | :D | Definite | <DET><N:fsD> | معرّف | الرسالة |
| | :a | construct state | <N:msaN><DET><N:msDG> | مضاف | مقعدُ الرجلِ |
| ADV | :i | indefinite | <N:fsiN> | نكرة | مقعدٌ |
| <N>,<A> | Case | | | | |
| | :N | Nominative | | مرفوع | رجلٌ |
| <ADV> | :A | Accusative | | منصوب | رجلاً |
| | :G | Genitive | | مجرور | رجلٍ |
| <V> | Voice, Aspect Mode | | | | |
| | :a | active | | معلوم | يَكتبُ |
| | :b | passive | | مجهول | يُكتَبُ |
| | :P | Perfect | <CONJC><V+nopro:aP3ms> | ماضٍ | وضربوا |
| | :I | Imperfect | <CONJS+subjunc><V+pro:aI3mp><PRO+acc:3fs> | مضارع | ليضربوها |
| | :Y | Imperative | <V+pro:Y3mp><PRO+acc:3mp> | أمر | إضربوهم |
| | :F | Active Participle | <V:FmsiA> | إسم فاعل | ضارباً |
| | :M | Passive Participle | <V:MfsiA> | إسم مفعول | مضروبةً |
| | :N | iNdicative | | مرفوع | |
| | :S | Subjunctive | | منصوب | |
| | :J | Jussive | | مجزوم | |
| | :E | Energetic | | مؤكد | |
| <V>,<PRO> | Person | | | | |
| | :1 | 1st person | | متكلّم | |
| | :2 | 2nd person | | مخاطب | |
| | :3 | 3rd person | | غائب | |

Table 5.1.c. Semantic and other syntactic features and values in Arabic-Unitex. Semantic encodings *in italics* in the table are not encoded systematically in the dictionary and depend on the requirements of a domain

| POS carrying the feature | Code | In English | Encoded examples | In Arabic | Arabic examples |
|---|---|---|---|---|---|
| \<N>\<PREP>\<PRO>\<PRTCL> | Case | | | | |
| | +Nom | Nominative | \<PRO+Ppers+Nom:1s> | مرفوع | أنَا |
| | +Acc | Accusative | \<PRO+Ppers+Acc:3d> | منصوب | ضربهما |
| | +Gen | Genitive | \<PREP+pro> \<PRO+Ppers+Gen:3d> | مجزوم | بهمَا |
| \<CONJS>\<PRO>\<PRTCL> | Mode | | | | |
| | +indic | Governs indicative | \<CONJS+indic+nopro> | مرفوع | قَدْ |
| | +subjunc | Governs subjunctive | \<CONJS+subjunc+nopro> | منصوب | لَنْ |
| | +juss | Governs jussive | \<CONJS+juss+nopro> | مجزوم | لَمْ |
| | | | | | |
| \<PREP>\<V>\<N>\<A> | +pro | form with mandatory enclitic | \<PREP+pro> \<PRO+Ppers+Gen:3fs> | | بها |
| \<PREP>\<V>\<N>\<A> | +nopro | form incompatible with enclitic | \<V+nopro:aP3mp> | | كتبوا |
| | | | | | |
| \<N>\<A> | *+Hum* | Human | | | طبيب |
| | *-Hum* | non-Human | | | دفتر |
| \<N>\<PREP>\<PRO> | *+Loc* | Locative | \<PRO+Pinterrog+Loc> | | أين؟ |
| \<N>\<PREP>\<PRO> | *+Temp* | Temporal | \<PREP+nopro+Temp> | | طِيلَة |
| \<PRTCL> | *+Vocative* | PRTCL | \<PRTCL+Vocative > | | يا أيُها |
| \<N> | *+Abst* | Abstract | \<N+Abst:ms> | | حصول |
| | | | \<N+Instance:fs> such as shippment | إسم مرة | شحنة |
| | *+generic* | | \<N+generic:ms> such as shipping | | شحن |
| \<N> | *+Anml* | Animal | \<N+Anml:ms> | | حصان |
| \<N> | *+AnmlColl* | collective animal | \<N+AnmlColl:fs> | | ماشية |
| \<N> | *+Conc* | Concrete | \<N+Conc:fs> | | طاولة |
| \<N> | *+ConcColl* | collective concrete | \<N+ConcColl:p> | | بهارات |
| \<N> | *+HumColl* | Collective | \<N+HumColl:msiN:ms> | إسم جمع | شعب |
| | *+ species* | Species | \<N+AnmlColl+species:ms> | اسم جنس جمعي | **بقر** |
| | *+count* | countable species | \<N+Anml+count:fs> | إسم الواحد | **بقرة** |
| | *+uncount* | Uncountable | \<N+Anml+uncount:fs> | **اسم الجنس** | **لبن** |
| \<V> | *+t* | Transitive | \<V+t> | متعدّي | ضرب |
| \<V> | *+i* | Intransitive | \<V+i> | لازم | جاء |
| \<V>, \<N>, \<A>,\<ADV> | *+z1* | General vocab. | \<N+z1> | مفردات عامة | دفتر |
| \<V>, \<N>, \<A>,\<ADV> | *+z2* | Specialized vocab. | \<N+z2> | مفردات متخصصة | برنت – بربون |
| \<V>, \<N>, \<A>,\<ADV> | *+z3* | very specialized | \<N+z3> | متخصصة جداً | الإيكسيتون |